\documentclass[a4paper,fleqn, review]{cas-sc}

\usepackage{amsmath}

\usepackage[numbers, compress]{natbib}

\usepackage{siunitx}
\usepackage{soul}

\usepackage{tabularx}
\usepackage{float}
\usepackage{graphicx}
\graphicspath{{../../Images/}}
\usepackage{subfig}
\usepackage{amsmath}
\usepackage{epstopdf} 
\usepackage{hyperref}
\def\tsc#1{\csdef{#1}{\textsc{\lowercase{#1}}\xspace}}
\tsc{WGM}
\tsc{QE}
\tsc{EP}
\tsc{PMS}
\tsc{BEC}
\tsc{DE}

\begin{document}

\title{Evaluating AI capabilities in detecting conspiracy theories on YouTube}
\author[1]{Leonardo La Rocca}
\author[1,2]{Francesco Corso}
\author[1]{Francesco Pierri}

\address[1]{Dipartimento di Elettronica, Informazione e Bioingegneria, Politecnico di Milano,Via Ponzio 34, Milan, Italy}
\address[2]{CENTAI,Corso Inghilterra 4, Turin, Italy}

\begin{abstract}
As a leading online platform with a vast global audience, YouTube's extensive reach also makes it susceptible to hosting harmful content, including disinformation and conspiracy theories.
This study explores the use of open-weight Large Language Models (LLMs), both text-only and multimodal, for identifying conspiracy theory videos shared on YouTube. 
Leveraging a labeled dataset of thousands of videos, we evaluate a variety of LLMs in a zero-shot setting and compare their performance to a fine-tuned RoBERTa baseline. Results show that text-based LLMs achieve high recall but lower precision, leading to increased false positives. Multimodal models lag behind their text-only counterparts, indicating limited benefits from visual data integration. To assess real-world applicability, we evaluate the most accurate models on an unlabeled dataset, finding that RoBERTa achieves performance close to LLMs with a larger number of parameters. Our work highlights the strengths and limitations of current LLM-based approaches for online harmful content detection, emphasizing the need for more precise and robust systems.
\end{abstract}


\maketitle

\section{Introduction}
\label{sec:introduction}


In the digital era, social media platforms have become integral to modern communication, serving not only as entertainment hubs but also as primary sources of information. YouTube, in particular, plays a crucial role in shaping public discourse, with millions of videos uploaded daily covering diverse topics, including news, science, and politics \cite{youtube_as_public_sphere, youtube_daily_videos}. 
However, the increasing reliance on social media as an informational platform has also amplified the spread of harmful content. This phenomenon presents multiple societal risks, including the reinforcement of extremist ideologies, the proliferation of prejudices, and, in some cases, the incitement of violence.
Conspiracy theories represent a specific and pervasive form of harmful content. These narratives typically challenge established knowledge and institutional credibility, often relying on anecdotal evidence, logical fallacies, and selective framing of facts \cite{how_cts_construct_oppositional_videos_on_yt_Grusauskaite_2022}. 
A defining characteristic of conspiracy theories is their self-sealing nature: any counter-evidence is often dismissed as part of the alleged conspiracy, making them particularly resilient to fact-checking and rational discourse.
The real-world impact of conspiracy theories has been well documented.
For example, anti-vaccine narratives gained significant traction during the COVID-19 pandemic, contributing to vaccine hesitancy and hampering public health efforts \cite{vaccine_hesitancy_pierri_2022}. 
Similarly, the QAnon conspiracy theory, which propagated claims of a secret elite cabal, played a role in radicalizing individuals, culminating in the January 6th Capitol Hill attack in the United States \cite{capitol_hill_attacks}.
These cases highlight the tangible societal risks posed by conspiracy content when left unchecked.
In response to the growing concerns around misinformation, YouTube has introduced various moderation policies aimed at curbing the spread of harmful content.
However, content moderation at YouTube's scale remains an immense challenge, as manual review processes are insufficient to address the sheer volume of uploaded content \cite{google_misinformation}.
Advances in artificial intelligence (AI) and natural language processing (NLP) offer promising solutions for content moderation, enabling scalable detection of problematic narratives \cite{detection_and_moderation_of_detrimental_content_on_sm_Gongane2022}.
Large Language Models (LLMs) offer high-quality zero-shot classification, enabling accurate annotation even in settings with limited labeled data, and their multimodal capabilities align well with YouTube’s content format. 
Recent work has shown that LLMs not only rival fine-tuned baselines \cite{ct_tiktok_Corso_Pierri_Morales_2024} but can also outperform human annotators in detecting harmful or conspiratorial content \cite{harmful_yt_video_detection_Wonjeong_2024, MUSE_Zhou_Sharma_Zhang_Althoff_2024}, demonstrating their promise for automated content moderation.
This study explores the development of classification systems based on Large Language Models (LLM) to identify conspiracy content on YouTube, aiming to enhance automated moderation efforts and contribute to a safer information ecosystem.

Specifically we seek to answer the following research questions:

\begin{itemize}
    \item \textbf{RQ1:} How effectively do modern open-weight LLMs detect conspiracy content on YouTube using only textual data, such as transcripts, descriptions, and comments?
    
    \item \textbf{RQ2:} Does incorporating visual information in multimodal LLMs significantly enhance their ability to identify conspiracy content compared to text-only models?
\end{itemize}

To provide answers to these research questions, we leverage YouNiCon, a manually labeled dataset of conspiratorial YouTube videos~\cite{younicon_liaw_2023}. 
We then enrich this dataset with additional metadata, including textual elements (transcripts, titles, descriptions, comments) and image data (thumbnails, frames), retrieved directly from YouTube.
For our evaluation of the LLM performance, we evaluate a range of different models: 
We begin by including smaller LLMs (7-8 billion parameters), such as \texttt{Llama-3.1-8B-Instruct}, \texttt{Mistral-7B-Instruct-v0.2}, and \texttt{Gemma-1.1-7B-IT}, alongside a larger model, \texttt{Llama-3.3-70B-Instruct}. To specifically address \textbf{RQ2}, we also use \texttt{Llama-3.2-11B-Vision-Instruct}, a small multimodal LLM (11 billion parameters) that processes both text and a single image.


    
    

We prompted each model using two strategies to examine the impact of prompt design on classification accuracy. 
Additionally, we evaluate ensemble models, combining multiple LLMs to enhance prediction robustness. 
These results are benchmarked against a baseline RoBERTa model, fine-tuned for text classification using the labeled dataset.

Following the evaluation on the labeled dataset, we apply the best-performing models to an unlabeled sample of approximately \num{9000} YouTube videos. 
This experiment provides insights into the broader distribution and prevalence of conspiracy content on the platform. 
A subset of these model-generated labels is then manually annotated to assess model performance in real-world settings.

By exploring multiple models, prompting techniques, and real-world applications, this research aims to provide a comprehensive evaluation of LLM capabilities in conspiracy content detection and contribute to the development of scalable, automated moderation tools for detecting conspiracy content on YouTube, informing future advancements in AI-driven content moderation.

The paper is structured as follows. 
In Section 2 we discuss meaningful related work to the current study, especially on harmful content detection, large language models (LLMs), and multimodal approaches to content classification. 
Section 3 outlines the methodology, detailing the dataset collection process, experimental setup, evaluation metrics, and the in-the-wild analysis of conspiracy content classification. 
Section 4 presents the results, including performance evaluation through precision-recall metrics and Normalized Expected Cost (NEC). 
In this section we further explore the impact of vision models, prompting strategies, and ensemble methods, along with an analysis of real-world labeling performance. 
Finally, in Section 5 we summarize its contributions, discussing its implications and limitations, and outlining ethical considerations and future research directions.

\section{Related Work}

Recent research on conspiracy content detection and misinformation analysis has employed various classification methodologies, primarily using BERT-based models, machine learning classifiers, and LLMs. The PAN at CLEF 2024 task on conspiracy theory classification \cite{pan_at_clef_24} reinforced the effectiveness of BERT-based models. CT-BERT \cite{CT-BERT}, a domain-specific variant, was the state-of-the-art model, demonstrating high performance in binary and fine-grained classification tasks. Other models of the BERT family, such as RoBERTa \cite{roberta}, also proved reliable, achieving F1 scores of ~0.9. 

Beyond transformer models, traditional machine learning and deep learning approaches have also been explored. \citet{longitudinal_analysis_yt_recommendation_faddoul_2020} analyzed YouTube’s recommendation system using a machine learning classifier trained on metadata, transcripts, and comments. Their multi-stage pipeline leveraged fastText \cite{fastText_Joulin_2017} embeddings and logistic regression, weighting comment-based signals highest (52\%) in classification decisions. The model, validated via 100-fold cross-validation, achieved 78\% precision and 86\% recall. Similarly, \citet{just_a_flu_Papadamou_2021} developed a deep learning classifier to detect pseudoscientific content, outperforming traditional models like SVM and Random Forest. By integrating multiple textual inputs, their approach provided a more comprehensive understanding of misinformation patterns while also highlighting the correlation between conspiracy content and high user engagement.

More recent studies have focused on LLM-based approaches, investigating their potential for automated content moderation. \citet{ct_tiktok_Corso_Pierri_Morales_2024} examined Llama 3 (8B), Mistral (7B), and Gemma (7B) \cite{meta_llama_2024, mistral_7B_2023, google_gemma_7B_2024} for conspiracy detection on TikTok, using OpenAI Whisper \cite{whisper_2022} for audio transcription. They tested multiple prompting strategies, finding that classification performance varied significantly based on input type and prompt formulation. Compared to a fine-tuned RoBERTa baseline (0.83 precision/recall in a balanced setting, 0.68/0.67 in an unbalanced setting), LLMs showed inconsistent results, emphasizing the challenge of optimizing prompt-based classification. Expanding beyond text-based moderation, \citet{MUSE_Zhou_Sharma_Zhang_Althoff_2024} introduced MUSE, a retrieval-augmented LLM for misinformation correction, integrating real-time web search to verify claims and assess credibility. Evaluated on 247 tweets with embedded YouTube videos, MUSE outperformed GPT-4 by 37\% in identifying inaccuracies and providing evidence-based corrections. This study also demonstrated MUSE’s effectiveness in detecting multimodal misinformation, such as misleading image-text pairings, underscoring the value of retrieval-augmented models in combating evolving misinformation. Additionally, \citet{harmful_yt_video_detection_Wonjeong_2024} applied GPT-4-Turbo to detect harmful YouTube content, leveraging a dataset of 19,422 videos. Their model incorporated text metadata and visual elements, improving classification accuracy and significantly outperforming human crowdworkers, reinforcing the potential of multimodal LLMs for scalable content moderation.

These studies highlight the evolution of classification methods, from traditional machine learning to advanced LLMs, each presenting unique strengths and challenges in conspiracy content detection. So far, foundational transformer-based models such as BERT appear to be the most effective. Yet, a more profound and continuous analysis of novel LLMs and multimodal LLMs is necessary to assess the evolution of the state-of-the-art. In this context, \citet{younicon_liaw_2023} introduced YouNICon, a curated dataset of YouTube videos from channels flagged as conspiracy-related by Recfluence. The dataset includes metadata from 1,912 labeled channels, filtering for English transcripts to compile 596,967 videos. To create labeled data, 2,200 videos were annotated by Amazon Mechanical Turk (AMT) crowdworkers, and a fine-tuned BERT model generated pseudo-labels for the remaining dataset. A further refinement process manually annotated an additional 1,000 videos, resulting in a final labeled subset of 3,161 videos. RoBERTa-large was used to evaluate classification performance, achieving 0.62 precision and 0.75 recall. This dataset is crucial for studying conspiracy content dissemination and was used in our study to assess model performance.

\section{Data and Methods}

\subsection{Datasets}

To perform our study, we leveraged the YouNICon dataset \cite{younicon_liaw_2023}: a manually labeled dataset of conspiracy videos on YouTube (N=3161).
In addition to video IDs and class labels, YouNICon provides other metadata, such as video titles, descriptions, and transcriptions. However, to fully leverage LLM capabilities, we required a more extensive set of features, including channel metadata and images from the videos themselves.
Therefore, while YouNICon served as the foundation for this research, it was necessary to enrich it with additional data to better capture the contextual signals present in YouTube videos.

To maximize the amount of relevant information provided to the models, we retrieved multiple sources of textual and visual data for each video.
Transcripts were a primary focus, as they provided a textual representation of the video content. Both auto-generated transcripts and user-uploaded transcripts (when available) were collected to ensure the highest quality of speech-to-text data. 
Additionally, we extracted channel metadata, which included information such as the channel's name and subscriber count, features that might provide contextual clues about the credibility of the content.
We also extracted user comments, which can serve as valuable indicators of how content is perceived and engaged with by viewers.
Given the vast number of comments on some videos, it was impractical to retrieve all comments. 
Instead, the collection process was constrained to a maximum of 20 comments per video, with up to five replies per comment, these comments were sorted using YouTube’s “Top” ranking system.
Furthermore, we extracted the thumbnail of the video, a random selection of frames (N=3) and a uniformly sampled sequence of frames (N=3) from each video.
These images were included to support the evaluation of the multimodal LLM.


We relied on \texttt{yt-dlp} \cite{yt_dlp} to retrieve transcripts, comments, thumbnails, and even full video files. Once the videos were downloaded, we extracted frames using the \texttt{ffmpeg} library \cite{ffmpeg}.

\begin{figure}[!t]
    \centering
    \includegraphics[width=.75\linewidth]{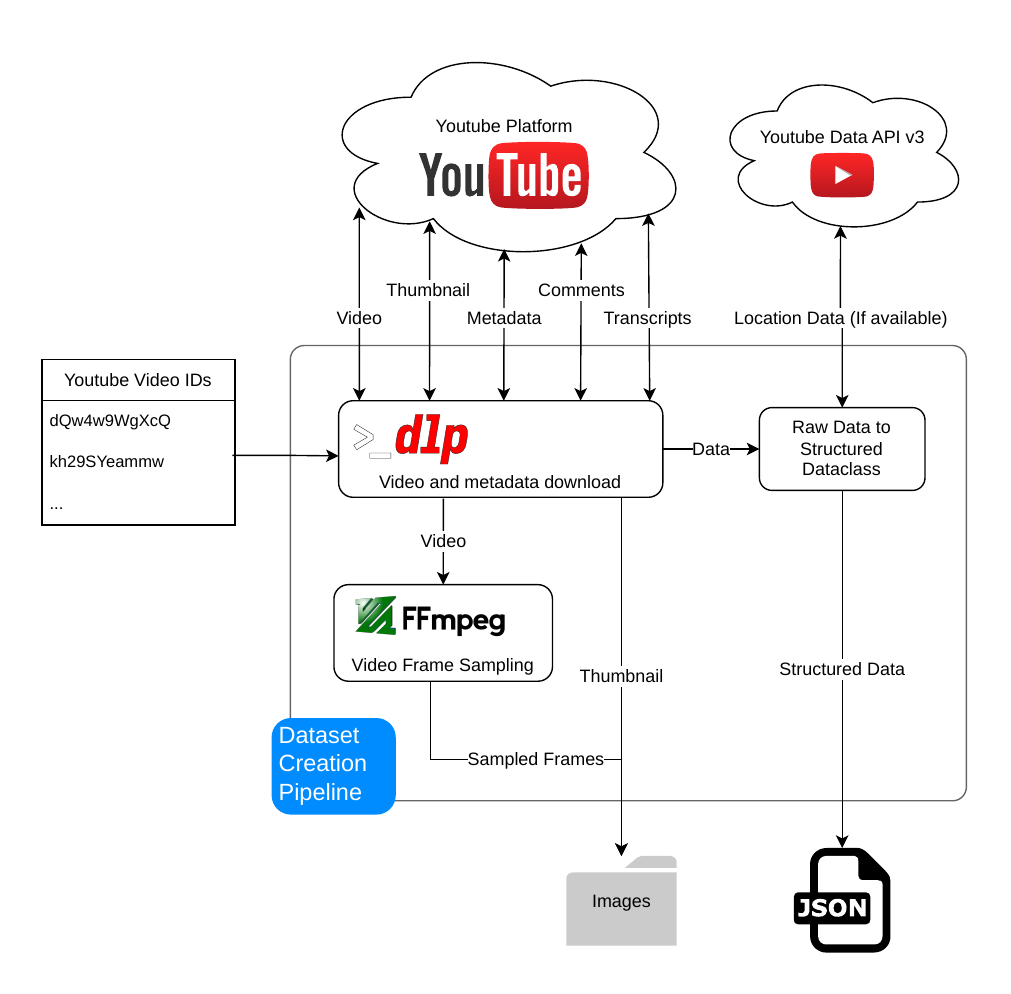}
    \caption{The dataset creation pipeline.}
    \label{fig:hydration pipeline}
\end{figure}



One challenge encountered during the data collection process was video availability: some videos listed in YouNICon had been removed by YouTube or made private by their creators. 
We excluded these videos from further analysis.




After filtering out unavailable videos, the final dataset consisted of 2,515 videos, with 897 labeled conspiracy-related (positive class) and 1,618 labeled as non related to conspiracy theories (negative class). 
This distribution, while unbalanced, provided a sufficient number of positive samples for evaluation.




We created a second dataset to validate the effectiveness of the best-performing models in-the-wild.
The goal was to assess how well the models generalize beyond the curated YouNICon dataset and whether they maintain consistent performance when exposed to real-world, naturally occurring content.
Since the YouNICon dataset was manually labeled and selected to include as much conspiracy content as possible, it did not necessarily reflect the broader distribution of content found on YouTube.




Thus, we identified a publicly available dataset containing approximately 2 million YouTube videos: Youtube-Commons~\cite{youtube_commons}.
The YouTube-Commons dataset is the largest corpus of freely licensed YouTube videos, comprising approximately 2 million videos released under the CC-By license, ensuring that they can be freely accessed and analyzed without violating copyright restrictions. 
While this dataset contained valuable metadata, it lacked several key features required for model evaluation, such as video descriptions and user comments.
Moreover, our primary interest was in obtaining video IDs that could be used for random sampling.
To address this, we extracted a random sample of 10,000 video ids from the dataset and proceeded to retrieve the missing information using the same yt-dlp-based pipeline that was previously employed for hydrating the YouNICon dataset. 
This process allowed us to construct a dataset with the same structure and features as our earlier experiments, ensuring consistency in evaluation and facilitating direct comparisons between results.

During the retrieval process, a significant number of videos (N=782) were found to be unavailable, either due to deletion, privacy restrictions, or removal by YouTube’s moderation policies. 
After filtering out unavailable videos, we obtained a final dataset of 9,218 instances, which served as the basis for our in-the-wild classification experiments.
From these set of videos we removed the ones that were in non-English languages using the \textit{langdetect} library in Python \cite{langdetect_pypi}, a port of the original Java library developed by Nakatani Shuyo \cite{nakatani_2010_langdetect}.


To assess the performance of the models in real-world scenarios, we conducted a manual annotation of a subset of the automatically labeled videos. Specifically, we selected videos where the models produced conflicting classifications. 
Since the models agreed on the remaining cases, evaluating their relative performance required human annotations for the disagreements to determine which model was more accurate.

In total, there were 125 videos with conflicting labels identified as English based on their metadata.
However, during the annotation process, we found that some of these videos were actually in other languages despite having English titles.
To ensure the validity of the evaluation, we filtered out all videos where at least one annotator marked them as non-English, resulting in a final set of 87 videos.

The selected videos were labeled by two annotators, with a third annotator resolving cases where the initial two disagreed.
After labeling, we measured inter-rater reliability using Cohen’s kappa \cite{Cohen1960_kappa}, a statistical measure that accounts for agreement occurring by chance. 
The computed Cohen’s kappa for the filtered dataset was 0.33, indicating fair agreement between the annotators. 
This result highlights the inherent subjectivity of identifying conspiracy content, as some cases may fall into a gray area where different annotators interpret the narratives differently.

\subsection{Models}

The study was structured to compare the performance of text-only LLMs and a multimodal model, examining whether visual information contributes meaningfully to classification. To ensure a controlled evaluation, we standardized prompting strategies across models. Additionally, a RoBERTa model was fine-tuned as a baseline to compare against the LLM-based classifiers.

To conduct the experiments, we utilized the Hugging Face serverless inference API \cite{hf_inference_api}, which enabled inference on state-of-the-art LLMs without requiring dedicated local hardware.
The model selection process was driven by two key factors: availability through the API and comparability in terms of model size to ensure a meaningful performance analysis.
The text-only models chosen for evaluation were: \textbf{Mistral 7B} \cite{mistral_7B_2023}: version 0.2, \textbf{Google Gemma 7B} \cite{google_gemma_7B_2024}: version 1.1, and \textbf{Llama 3.1 8B} \cite{meta_llama_2024}. 
All these models are fine-tuned for instructions.
Additionally, we incorporated a larger LLM to evaluate whether scaling model size significantly improved classification performance: \textbf{LLaMA 3.3 70B} \cite{meta_llama_3_3_2024}: fine-tuned for instructions.

To assess the impact of incorporating visual data, we selected a multimodal model with a comparable parameter size: \textbf{LLaMA 3.2 11B} \cite{meta_llama_3_2_2024}, also fine-tuned for instructions.
Using a multimodal model of similar scale allowed us to isolate and measure the effect of visual input, particularly the addition of a single image (in this case, the video thumbnail) without confounding the results with differences in model size.

This set of models provided a comprehensive comparison across text-only, multimodal, and larger-scale LLMs, enabling a thorough evaluation of their respective capabilities in conspiracy classification.

We employed zero-shot prompting, meaning that the models were directly evaluated on the classification task without seeing any labeled examples.
The prompts used in the experiments were adapted from \citet{ct_tiktok_Corso_Pierri_Morales_2024}, which investigated the detection of conspiracy content on TikTok.
By using the same prompts as in their study, we aimed to reduce the number of experimental variables and enable more direct comparisons between results.

Two prompts were tested:

\begin{itemize}
    \item A \textbf{simple classification prompt} instructing the model to determine whether a given YouTube video contained conspiracy-related content: "\textit{Decide whether the following transcription of a video talks about a conspiracy theory or not (if yes output = 1/else output=0). Provide just your output, no justification.}".
    \item A \textbf{definition-based prompt} that included a formal description of a conspiracy theory \cite{douglas2023conspiracy} before requesting classification: "\textit{A conspiracy theory is a belief that two or more actors have coordinated in secret to achieve an outcome and that their conspiracy is of public interest but not public knowledge. Conspiracy theories (a) are oppositional, which means they oppose publicly accepted understandings of events; (b) describe malevolent or forbidden acts; (c) ascribe agency to individuals and groups rather than to impersonal or systemic forces; (d) are epistemically risky, meaning that though they are not necessarily false or implausible, taken collectively they are more prone to falsity than other types of belief; and (e) are social constructs that are not merely adopted by individuals but are shared with social objectives in mind, and they have the potential not only to represent and interpret reality but also to fashion new social realities. Given this definition of conspiracy theory: Decide whether the following transcription of a video talks about a conspiracy theory or not (if yes output = 1/else output=0). Provide just your output, no justification.}".
\end{itemize}

To standardize outputs, the models were instructed to return only the classification label without any accompanying reasoning or explanation.
Given the constraints imposed by LLM context windows, careful selection of input features was necessary. 
The textual input provided to each model consisted of a combination of metadata, transcripts, and user comments, ensuring a comprehensive representation of video content. 

\begin{figure}[!t]
    \centering
    \includegraphics[width=\linewidth]{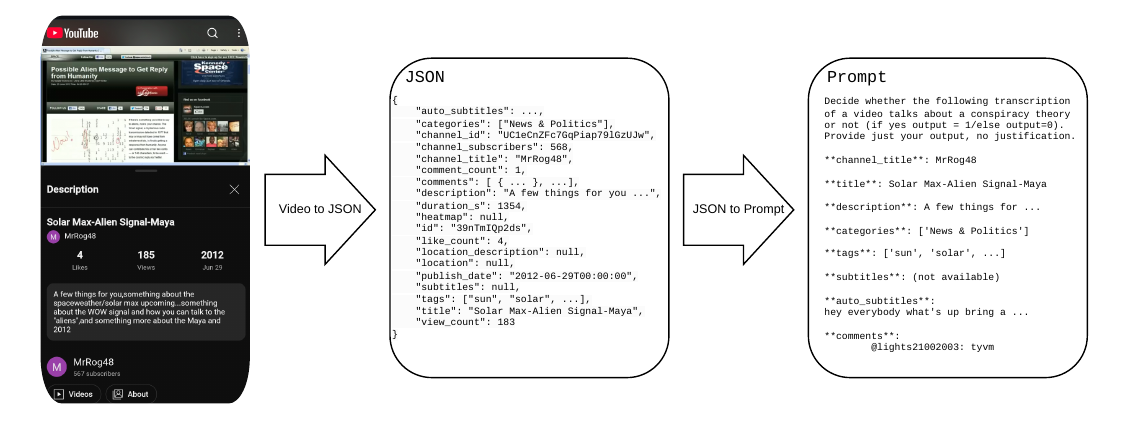}
    \caption{From Youtube Video to Prompt.}
    \label{fig:video-to-prompt}
\end{figure}

For the multimodal model, the input was identical to the text-only models, with the addition of a single image, specifically the video thumbnail.


In addition to evaluating each model independently, we explored the potential benefits of ensemble classification, combining the outputs of all the smaller text-only LLMs.
By combining outputs from multiple LLMs, we sought to mitigate errors arising from individual model biases and evaluate whether ensemble predictions were more consistent and robust than single-model outputs.

To establish a baseline for comparison, we fine-tuned a RoBERTa model \cite{roberta} on the YouNICon dataset using a train-validation-test split of 0.75, 0.075, and 0.175, respectively. 
This allowed for direct assessment of how a supervised classifier trained on labeled examples performed relative to the zero-shot LLMs.

The model was trained using early stopping, a technique that monitors validation loss and halts training when further improvements are unlikely, reducing the risk of overfitting. 
The final evaluation was conducted on the unseen test set, ensuring that the reported performance metrics reflected the model’s ability to generalize beyond the training data.
During this process, training stability proved to be an issue. 
The model exhibited high variance in loss across epochs, suggesting potential instability in the optimization process. 
This variability indicates that further hyperparameter tuning and experimentation with alternative training strategies may be necessary to achieve more consistent and reliable learning in future work.

\subsection{Evaluation Metrics}
\label{subsec:evaluation metrics}


The first step in our evaluation involved computing fundamental classification metrics for each model. 
Specifically, we measured precision and recall for the positive class (conspiracy content), which provide a basic indication of how well the models identified conspiracy videos.




In real-world applications, the cost of a false positive (wrongly flagging a non-conspiracy video) might be different from the cost of a false negative (failing to detect an actual conspiracy video).
To address this, we computed the expected cost (EC) and the normalized expected cost (NEC), which incorporate a predefined cost matrix to reflect the relative severity of different classification decisions.

The expected cost is defined as:

\begin{equation*}
EC = \frac{1}{N} \sum_{i=1}^{N} C(\hat{y}_i, y_i)    
\end{equation*}

Where: \( N \) is the total number of instances, \( \hat{y}_i \) is the predicted class for instance \( i \), \( y_i \) is the true class of instance \( i \), and \( C(\hat{y}_i, y_i) \) is the cost of classifying instance \( i \) as \( \hat{y}_i \) when its true label is \( y_i \).

Since EC depends on the chosen cost matrix, we further computed the normalized expected cost (NEC), which enables a comparative evaluation by normalizing EC against the performance of a dummy classifier. The dummy classifier always makes the same prediction for all instances, and the NEC is computed by comparing each model’s EC against the best dummy classifier (i.e., the one with the lowest cost among all possible dummy classifiers). NEC is defined as: 

$NEC = \frac{EC}{EC_{dummy}}$.

Considering that  \( EC_{dummy} \) is the expected cost of the best dummy classifier.

The NEC provides a clear baseline for evaluating model performance: If \( NEC < 1 \), the model performs better than the best dummy classifier else the dummy classifier performs better than the model.

An additional advantage of EC is that it can be mathematically rewritten as \cite{ferrer_2023}:

\[
EC = \sum_{c_i} \sum_{c_j} C_{c_ic_j} P_{c_i} R_{c_ic_j}
\]
Where \( C_{c_ic_j} \) represents the cost of predicting class \( c_j \) when the true class is \( c_i \), \( P_{c_i} \) is the empirical estimate of the prior probability of class \( c_i \) in the dataset, \( R_{c_ic_j} \) is the fraction of samples from class \( c_i \) that were classified as class \( c_j \).
This decomposition allows us to explore how the models would perform under different class distributions by adjusting \( P_{c_i} \). Since conspiracy content is relatively rare on YouTube, adjusting these priors enables a more realistic evaluation by simulating strong class imbalances.

To better understand the impact of different cost assumptions, we experimented with multiple cost matrices, adjusting the relative penalty of false positives and false negatives. 
This approach allowed us to analyze how model performance changed under different decision-making scenarios, which is particularly relevant for practical applications where the severity of classification mistakes depends on policy and platform objectives. 
The choice of an appropriate cost matrix is typically informed by domain expertise, as it must reflect the real-world consequences of false positives and false negatives. By experimenting with different cost matrices, we assessed the robustness of model rankings under varying conditions and identified models that performed well across multiple evaluation frameworks.


The final step of our evaluation involved selecting the best-performing models.
Given that the proportion of conspiracy content on YouTube is significantly lower than non-conspiracy content, a model that produces too many false positives would result in an excessive number of incorrect classifications. In large-scale applications, this would lead to either substantial manual verification costs or user dissatisfaction due to incorrect flagging of non-conspiracy content. 
Thus, our primary goal was to select models that minimized false positives while maintaining reasonable recall for conspiracy content.
We used an evaluation framework inspired by ranking systems, which provided a structured way to compare models based on their NEC in different scenarios.


\section{Results}
\label{sec:results}


\subsection{Precision and Recall}
\label{subsec:precision and recall}

\begin{figure}[!t]
    \centering
    \includegraphics[width=1\textwidth]{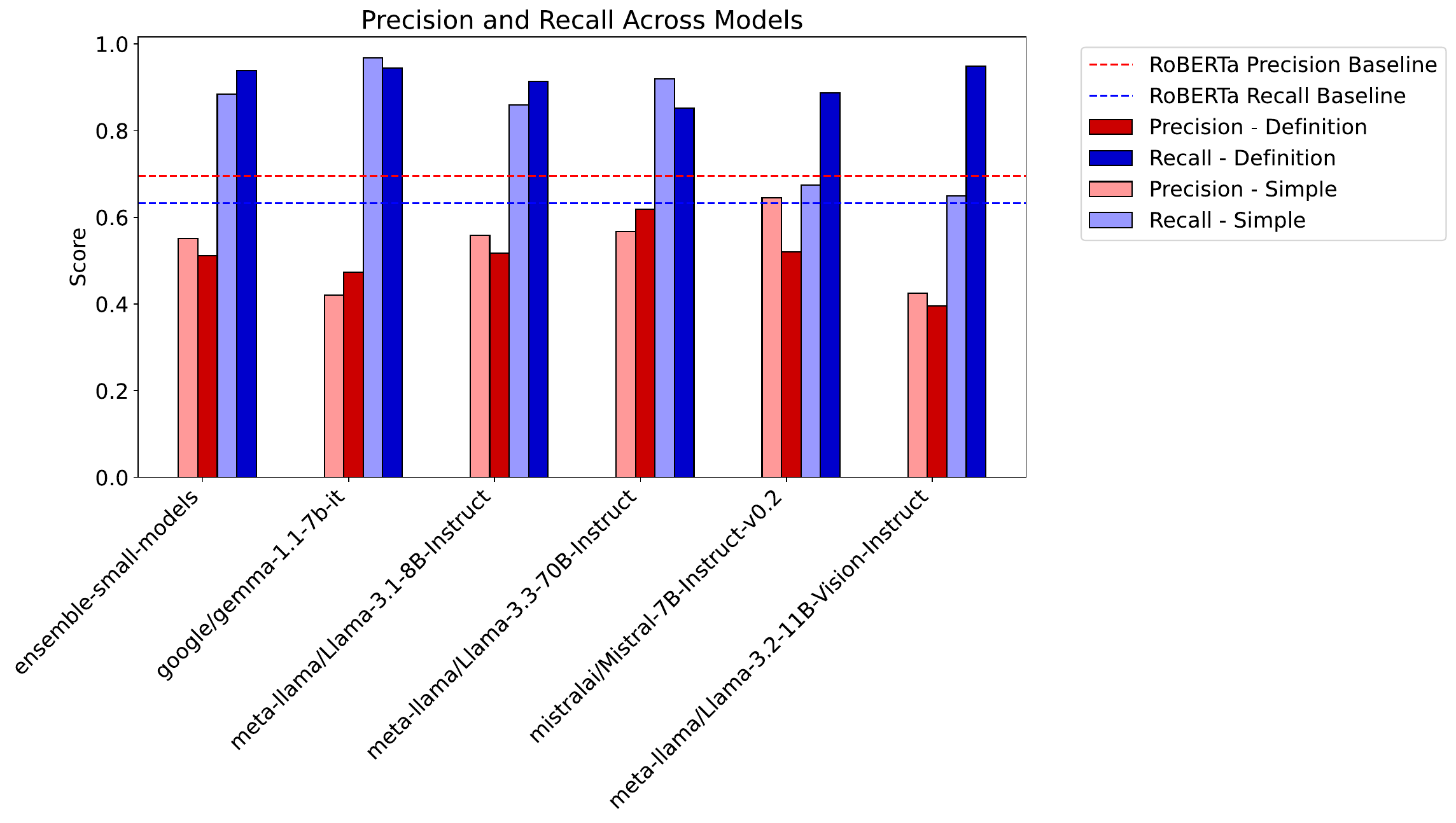}
    \caption{Precision and recall across models and prompts.}
    \label{fig:precision recall}
\end{figure}

\begin{table}[H]
\centering
\caption{Model performance: Precision and Recall.}
\begin{tabular}{l l c c}
\toprule
Model & Prompt & Precision & Recall \\
\midrule
Roberta 3Epochs & none & \textbf{0.695} & 0.633 \\
ensemble-small-models & simple & 0.551 & 0.884 \\
ensemble-small-models & definition & 0.512 & 0.939 \\
google/gemma-1.1-7b-it & definition & 0.474 & 0.944 \\
google/gemma-1.1-7b-it & simple & 0.421 & \textbf{0.968} \\
meta-llama/Llama-3.1-8B-Instruct & definition & 0.518 & 0.914 \\
meta-llama/Llama-3.1-8B-Instruct & simple & 0.559 & 0.859 \\
meta-llama/Llama-3.3-70B-Instruct & simple & 0.567 & 0.920 \\
meta-llama/Llama-3.3-70B-Instruct & definition & 0.618 & 0.852 \\
mistralai/Mistral-7B-Instruct-v0.2 & definition & 0.520 & 0.887 \\
mistralai/Mistral-7B-Instruct-v0.2 & simple & 0.645 & 0.674 \\
meta-llama/Llama-3.2-11B-Vision-Instruct & definition & 0.396 & 0.949 \\
meta-llama/Llama-3.2-11B-Vision-Instruct & simple & 0.425 & 0.649 \\
\bottomrule
\end{tabular}
\end{table}

In this section, we analyze precision and recall evaluated on the YouNICon dataset, which is unbalanced, containing 35\% positive labels and 65\% negative labels. 
Ignoring the RoBERTa baseline, a general trend observed across the models is a consistently high recall at the expense of precision. 
Models like \textit{google/gemma-1.1-7b-it} (simple prompt) exhibit extremely high recall (0.968), indicating a strong tendency to classify videos as conspiracy-related, even at the cost of increased false positives.
Conversely, models with higher precision, such as \textit{mistral-7B }(simple prompt), show a more balanced approach but at the expense of missing some true conspiracy videos.

This suggests that different models may be suitable for different applications: high-recall models are useful for exhaustive screening where false negatives must be minimized, while high-precision models are preferable in contexts where false positives must be tightly controlled. 
Identifying the best balance between these trade-offs remains crucial for effective deployment in real-world misinformation detection.
The model with the lowest recall is \textit{meta-llama/Llama-3.2-11B-Vision-Instruct }(simple prompt) at 0.649, indicating that it performs poorly in classifying conspiracy content, missing a significant number of true conspiracy videos while also maintaining a relatively low precision.



\textit{LLaMA 3.3 70B} (definition prompt) performs well, demonstrating strong recall and precision.
However, it is one order of magnitude larger in the number of parameters compared to other models, which could have implications for computational efficiency and deployment feasibility.

These results highlight the difficulty of striking a balance between recall and precision, where models optimized for one often suffer in the other, requiring careful selection based on the application’s needs. 
It is interesting to note that the general trend among these models is a high recall at the expense of precision. 
Hypothetically, this suggests the introduction of a bias during training that leans towards combating misinformation.
This bias may have been introduced by researchers for ethical considerations and/or to ensure compliance with ethical and legal requirements.

When balancing the dataset to ensure equal representation of conspiracy and non-conspiracy content, we observe an improved precision.
This is due to the effect of balancing the dataset, which reduces the instances of the negative class that was overrepresented in the imbalanced setting.
As a result, the number of true negatives and false positives decreases. 
Since true negatives do not impact precision or recall, the main effect is seen in false positives, which are inversely proportional to precision. 
Reducing false positives therefore results in an improvement in precision across all models.

\begin{figure}[!t]
    \centering
    \includegraphics[width=\textwidth]{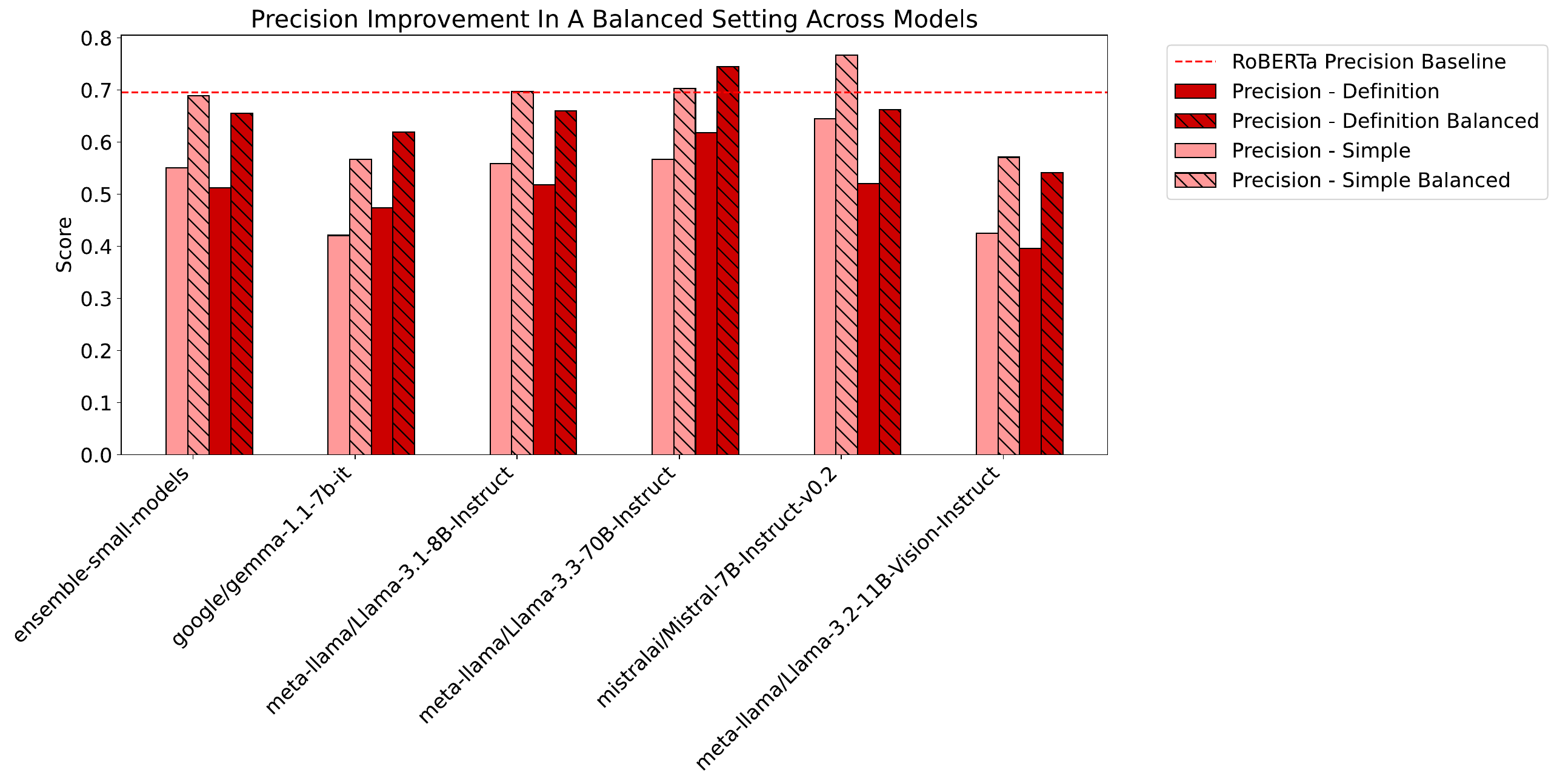}
    \caption{Precision in balanced dataset evaluation.}
    \label{fig:precision in balanced setting}
\end{figure}



\subsection{Normalized Expected Cost (NEC)}
\label{subsec:expected cost}

\begin{table}[H]
\centering
\caption{Model performance: NEC evaluated with the YouNICon class distribution. Lower is better.}
\begin{tabular}{l l c c c}
\toprule
Model & Prompt & NEC & Reduce FN & Reduce FP \\
\midrule
Always Positive & none & 1.804 & 1.000 & 16.234 \\
Always Negative & none & 1.000 & 4.989 & 1.000 \\
Roberta 3Epochs & none & 0.693 & 2.132 & \textbf{3.084} \\
ensemble-small-models & simple & 0.837 & 0.978 & 6.608 \\
ensemble-small-models & definition & 0.954 & 0.801 & 8.098 \\
google/gemma-1.1-7b-it & definition & 1.103 & 0.859 & 9.478 \\
google/gemma-1.1-7b-it & simple & 1.362 & 0.899 & 12.002 \\
meta-llama/Llama-3.1-8B-Instruct & definition & 0.938 & 0.900 & 7.751 \\
meta-llama/Llama-3.1-8B-Instruct & simple & 0.817 & 1.077 & 6.228 \\
meta-llama/Llama-3.3-70B-Instruct & simple & 0.784 & \textbf{0.790} & 6.411 \\
meta-llama/Llama-3.3-70B-Instruct & definition & \textbf{0.674} & 1.032 & 4.884 \\
mistralai/Mistral-7B-Instruct-v0.2 & definition & 0.931 & 1.015 & 7.477 \\
mistralai/Mistral-7B-Instruct-v0.2 & simple & 0.697 & 1.830 & 3.667 \\
meta-llama/Llama-3.2-11B-Vision-Instruct & definition & 1.496 & 1.053 & 13.061 \\
meta-llama/Llama-3.2-11B-Vision-Instruct & simple & 1.226 & 2.229 & 8.240 \\
\bottomrule
\end{tabular}
\end{table}

\begin{table}[H]
\centering
\caption{Cost Matrix}
\label{tab:cost_matrix}
\begin{tabular}{l c c c c}
\toprule
    Name & TN & FP & FN & TP \\
\midrule
    Basic & 0 & 0.5 & 0.5 & 0 \\
    Reduce FN & 0 & 0.1 & 0.9 & 0 \\
    Reduce FP & 0 & 0.9 & 0.1 & 0 \\
\bottomrule
\end{tabular}
\end{table}

The Normalized Expected Cost (NEC) metric provides a unified measure of misclassification penalties by incorporating false positives (FP) and false negatives (FN) into a single value.
By setting different costs for false negatives and false positives, we can evaluate models according to different needs. 
Additionally, a NEC less than 1 indicates that the model performs better than a dummy classifier, whereas a NEC greater than 1 suggests that the model performs worse than a dummy classifier.

Thus, we evaluate NEC on the YouNiCon dataset under three cost scenarios: equal penalties (basic), higher cost for false negatives (reducing FN), and higher cost for false positives (reducing FP).
The costs assigned to the various cases are resumed in Table \ref{tab:cost_matrix}.

\begin{figure}[!t]
    \centering
    \includegraphics[width=\textwidth]{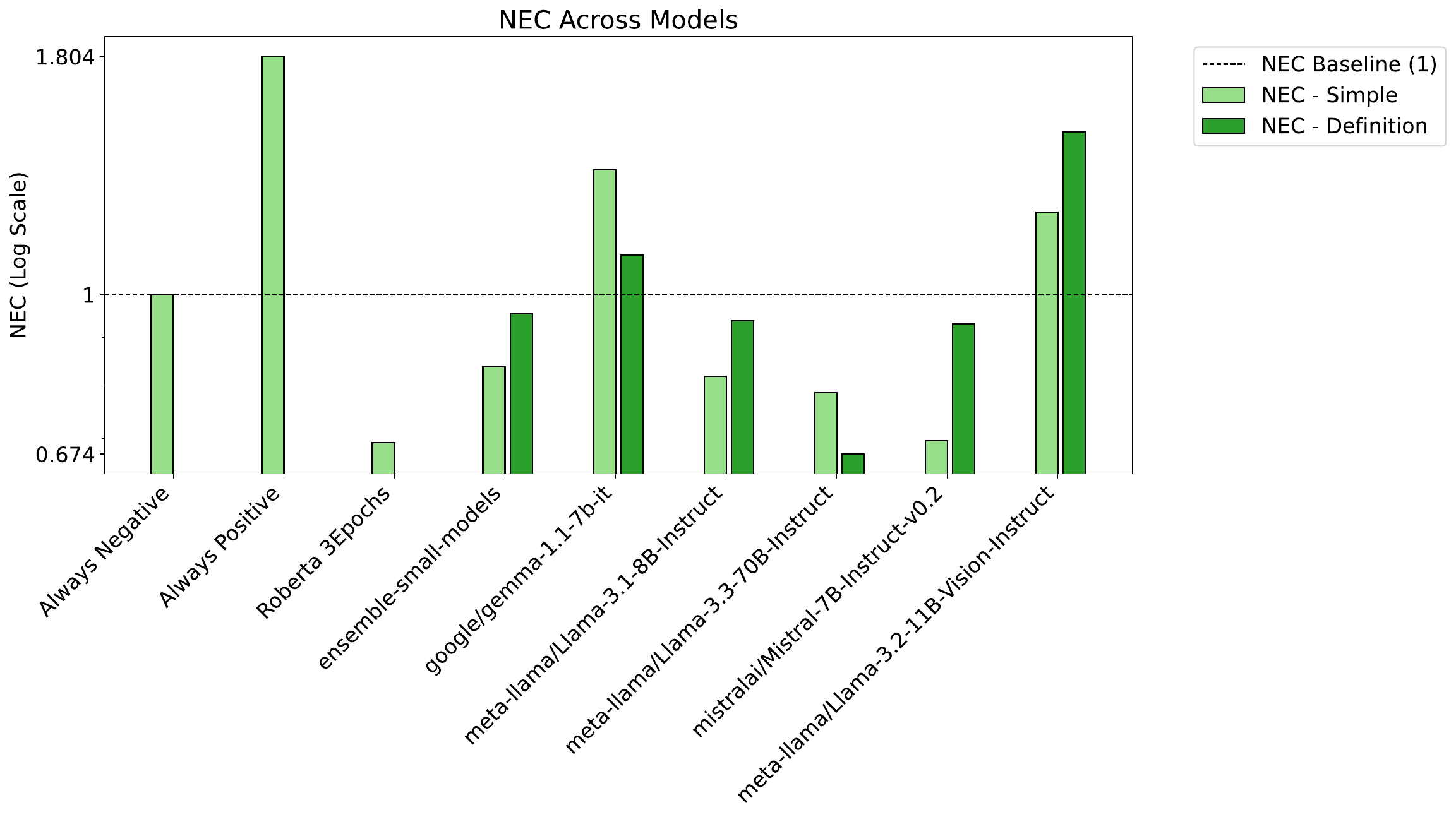}
    \caption{NEC across models and prompts.}
    \label{fig:NEC across models}
\end{figure}

In the \textbf{basic setting}, the best model appears to be the Llama 3.3 70B model with the definition prompt, followed by the Mistral 7B model with the simple prompt. Additionally, in this scenario, the ensemble model of the small LLMs performs well. 
On the other hand, the vision model and the Gemma model perform very poorly.

\begin{figure}[!t]
    \centering
    \includegraphics[width=\textwidth]{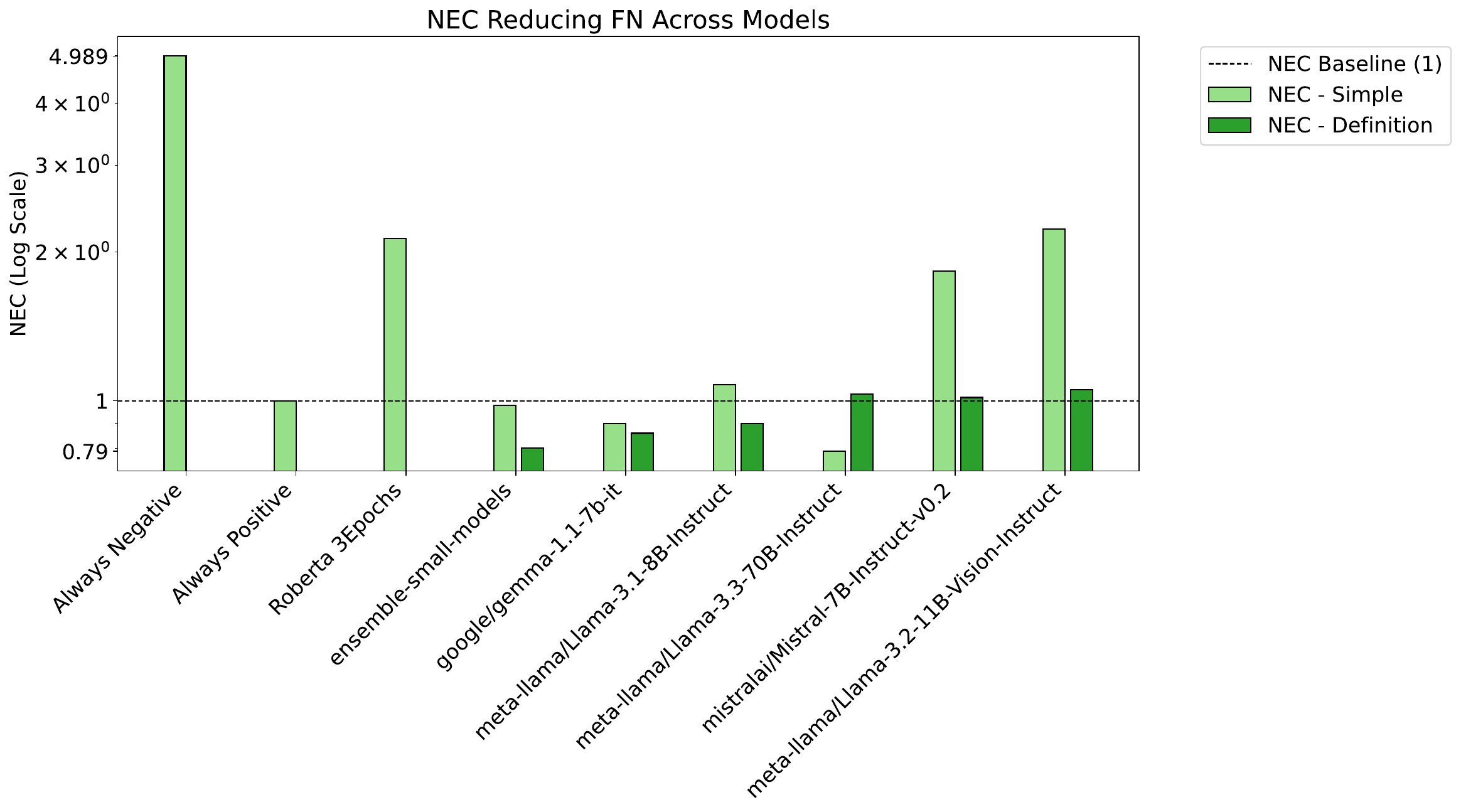}
    \caption{NEC reducing FN across models and prompts.}
    \label{fig:NEC reducing FN across models}
\end{figure}

In the \textbf{reducing FN} setting, the Gemma model demonstrates improved performance, suggesting a strong inclination towards minimizing false negatives. 
The ensemble models continue to perform well alongside the Llama 70B model, whereas the Mistral model encounters difficulties in this scenario.

\begin{figure}[!t]
    \centering
    \includegraphics[width=\textwidth]{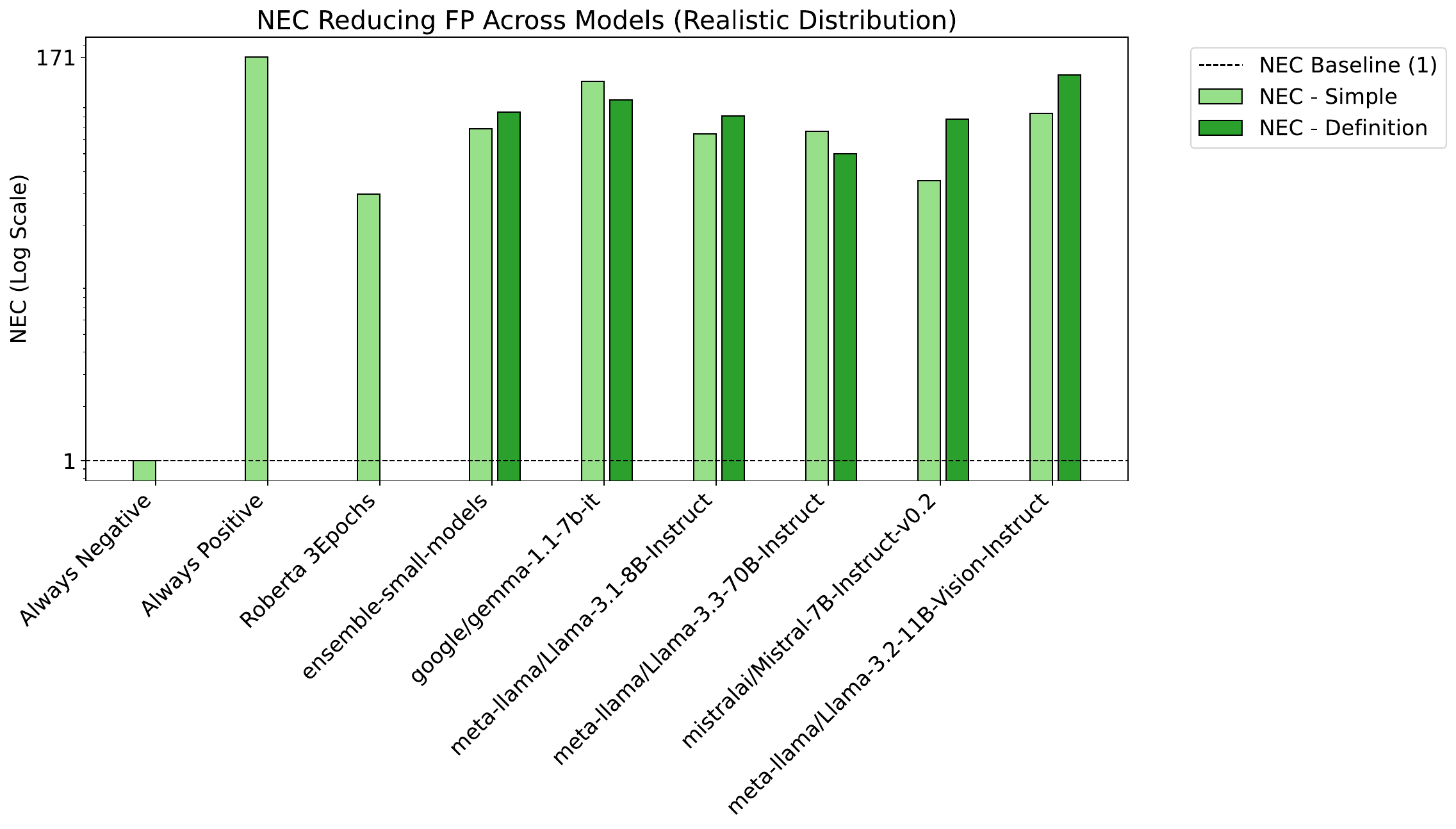}
    \caption{NEC reducing FP across models and prompts.}
    \label{fig:NEC reducing FP across models}
\end{figure}

In the \textbf{reducing FP} setting, all models struggle and perform poorly, with NEC values exceeding that of the dummy classifier. 
This suggests that the cost assigned to reducing false positives may be overly stringent. 
Additionally, it reinforces the observed model bias and highlights the inherent challenge in minimizing false positives. 
The impact of these high costs is further amplified by class imbalances in the YouNiCon dataset.

The general trend confirms what we observed in the precision and recall metrics: models perform well in both the basic setting and the reducing FN setting but struggle significantly when asked to reduce false positives.

Building on the properties of NEC discussed earlier, we extend our analysis to estimate model performance in a realistic scenario characterized by a strong class imbalance. In practice, NEC enables evaluation across any distribution, making it a flexible tool for assessing classifier performance. Based on prior research \cite{longitudinal_analysis_yt_recommendation_faddoul_2020}, we conducted experiments assuming a positive class presence of 5\%, which serves as an upper bound, as real-world distributions of conspiracy content are likely lower than 5\%. The results reaffirmed the trends observed in previous analyses, highlighting that models effectively minimize false negatives while continuing to struggle with reducing false positives. This evaluation provides insights into model behavior under deployment conditions, allowing for a more accurate assessment of their practical performance. Tables and graphs of the real-world setting evaluation can be found in the appendix \ref{sec:additional tables} \ref{sec:additional figures}.

\subsection{Discussion Of Results On YouNICon Dataset}
\label{subsec:gra}

\subsubsection{Performance of Vision Model}
\label{subsubsec:performance of vision model}

The vision model was one of the worst-performing models. 
This indicates that thumbnail images do not contribute meaningfully to classifying conspiracy content.
Furthermore, it shows that the inclusion of images may actually introduce noise, diminishing the model’s overall performance rather than enhancing it. 
This suggests that conspiracy content lacks distinctive visual signals, which limits the usefulness of image-based multimodal approaches. 
This hypothesis is further supported by the qualitative error analysis, where we found that a significant portion of conspiracy content is delivered in a “podcast" or “vlog" format, where the author talks in front of a camera without incorporating any graphics, videos, or presentations.
As a result, the thumbnail images may not provide meaningful classification cues and may instead suggest a neutral or normal context, reducing their effectiveness in distinguishing conspiracy-related content.

\subsubsection{Impact of Prompting Strategies}

Prompt design plays a crucial role in influencing model behavior and performance. 
The simple prompt performed better for smaller models with limited context size, while the definition prompt performed better for larger models with greater context size.
This suggests that larger models benefit from more explicit and structured prompts, allowing them to better utilize their extended context windows for nuanced reasoning and improved classification accuracy.
On the other hand, smaller models struggle with longer and more complex prompts, likely due to their restricted context capacity, making them perform better with concise and direct instructions.


\subsubsection{Performance of Ensemble Models}
\label{subsussec:performance of ensemble models}

Ensembling multiple models provides a robust approach to mitigating classification inconsistencies. The ensemble method helps smooth out individual model biases, leading to reduced variance and a more favorable NEC than some of the individual models. The ensemble composed of LLaMA 8B, Mistral 7B, and Gemma 8B, was evaluated on both the simple and definition prompt. The results show that while these ensembles performed better than some of the individual models, they did not provide a strong improvement overall, nor did they consistently surpass all the models they were composed of in performance. This indicates that ensembling, while useful in some cases, does not always lead to a clear advantage in detecting conspiracy content.

\subsubsection{Performance of RoBERTa model}

We use a fine-tuned RoBERTa model as a domain-specific baseline, demonstrating competitive performance in precision-recall trade-offs as demonstrated in other works in the literature \cite{la2023retrieving}. 
Overall, the RoBERTa model was the best-performing model, showing favorable results even when compared to the largest LLMs. 
However, it has a decisive advantage: it was fine-tuned on the labeled dataset that it was later evaluated on. Although the model had never seen the test set before, the fine-tuning process adapted it to the dataset and the annotators' biases, making it perfectly suited to the task needs.

The qualitative error analysis further reinforced this. Given the controversial nature of the assigned labels, models that understood dataset-specific nuances performed better. This raises a crucial question: would the RoBERTa model maintain its strong performance on an unseen dataset with different labeling biases and content characteristics? Additionally, it is important to note that the model was only trained in the imbalanced setting, meaning we do not have results for its performance in a balanced setting.

\subsection{In-The-Wild Experiment Results}
\label{subsec:in-the-wild results}

\begin{figure}[!t]
    \centering
    \includegraphics[width=\textwidth]{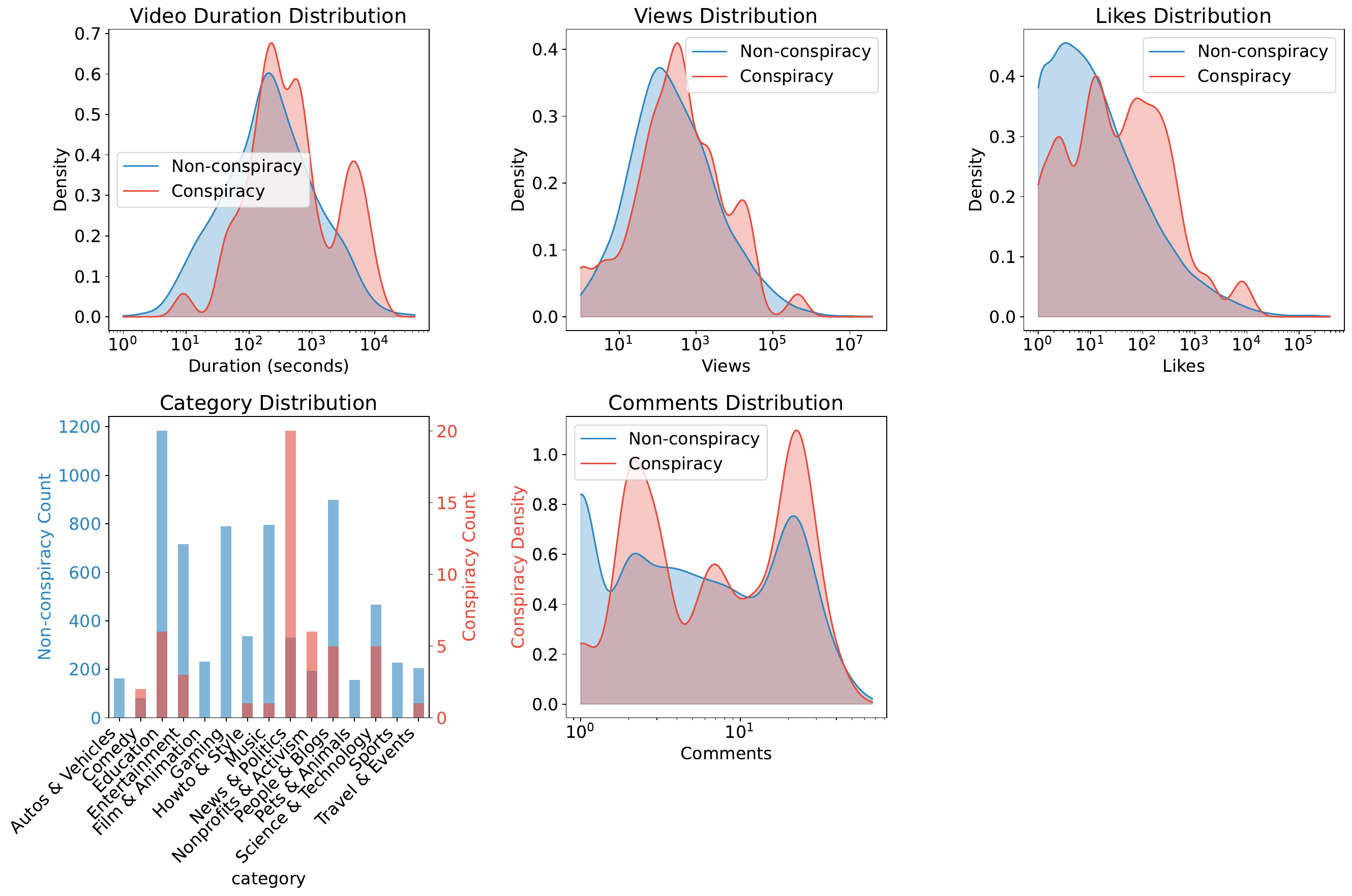}
    \caption{Distributions of the metrics od the in-the-wild dataset, divided by conspiracy and non conspiracy videos.}
    \label{fig:metrics}
\end{figure}

After evaluating model performance on the structured YouNICon dataset, we extended our analysis to an in-the-wild experiment, applying the best-performing models to a new dataset of 9,218 YouTube videos randomly sampled from the YouTube-Commons dataset. This phase aimed to assess how well models generalize beyond curated datasets in a real-world setting where the proportion of conspiracy content is naturally lower. After filtering the original dataset of 9,218 videos to exclude non-English content, we obtained a final dataset of 6,943 English videos. Within this dataset, the models identified 1.6\% of videos as conspiracy-related.
In Figure \ref{fig:metrics} we show the distributions of the metrics of the two classes identified by the models.
Notably, the videos classified as conspiratorial show to fall often in the “News and Politics category" and have more likes and views with regard to their non-conspiratorial counterpart, while also being averagely longer in duration.

To contextualize our findings, we performed a baseline analysis of the dataset and compared it with the distribution reported by \citet{dialing_for_videos_2023}. We leave more details about our evaluation in Appendix \ref{app:inw-dataset}.


To select the best models for in-the-wild experimentation, we ranked each model based on its NEC scores across multiple evaluation settings.
Given that different settings produced variations in rankings, we computed the median rank for each model to mitigate the influence of outlier rankings.
Based on this evaluation, RoBERTa, Mistral 7B (simple prompt), and LLaMA 3.3 70B (definition prompt) emerged as the best models.




The evaluation of model performance in the real-world scenario was conducted on a selectively labeled subset of videos, chosen based on classification disagreements between the Mistral and LLaMA models. Since only conflicting cases were annotated, the resulting metrics should not be interpreted as representative of overall model performance across the entire dataset. Instead, these numbers serve solely to compare the relative accuracy of the models in resolving ambiguous classifications. Consequently, any conclusions drawn from this analysis should be understood within the context of this targeted selection process.

\begin{table}[H]
\centering
\caption{Model performance: Precision and Recall.}
\begin{tabular}{l l c c}
\toprule
Model & Precision & Recall \\
\midrule
Llama & 0.429 & \textbf{0.800} \\
Mistral & 0.075 & 0.200 \\
RoBERTa & \textbf{0.455} & 0.333 \\
\bottomrule
\end{tabular}
\end{table}


\begin{figure}[!t]
    \centering
    \includegraphics[width=\textwidth]{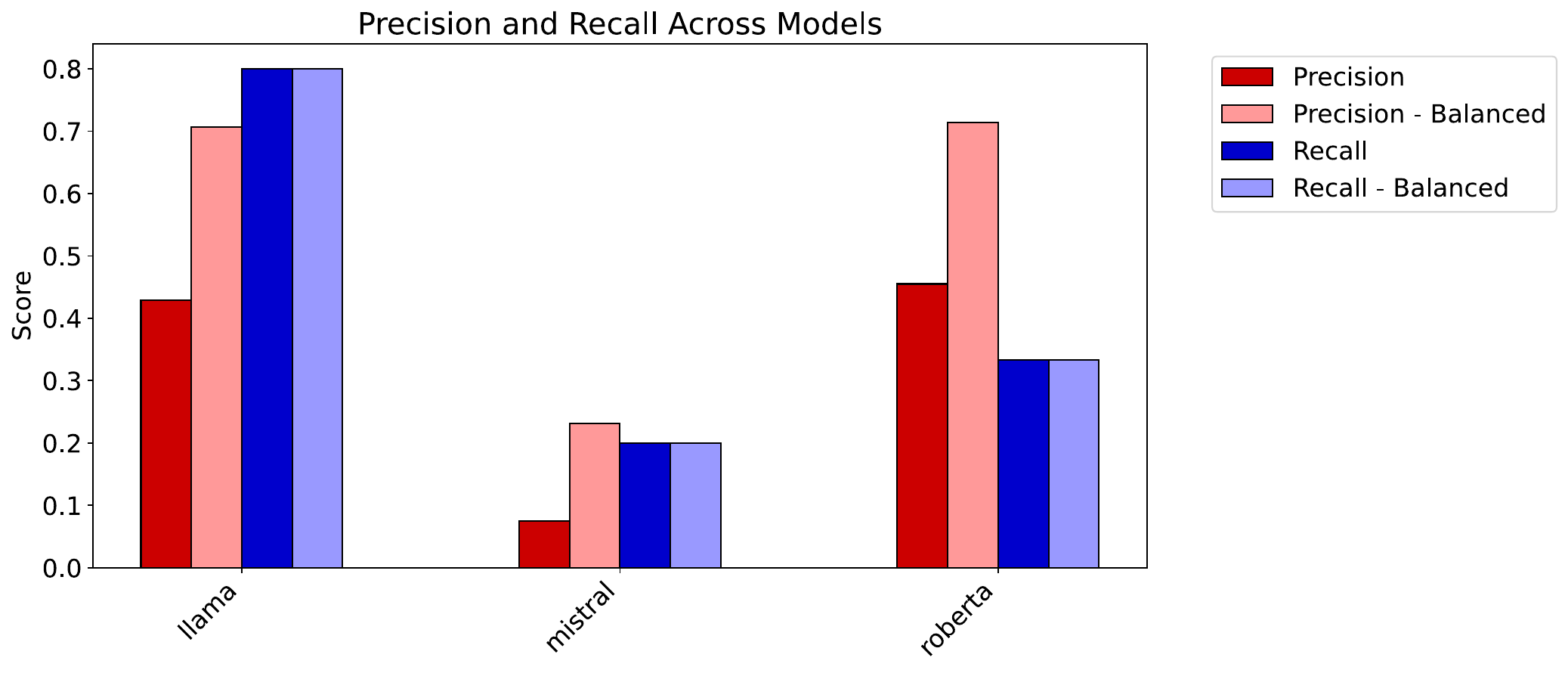}
    \caption{Precision and Recall across models.}
    \label{fig:precision_recall_itw}
\end{figure}

From the results, LLaMA outperforms Mistral, which is expected given its significantly larger parameter count (70B vs. 7B). Notably, RoBERTa achieves the highest precision among all models, making it particularly effective in minimizing false positives, an essential property for conservative classification approaches. Despite its lower recall, RoBERTa’s balance between precision and false positive reduction makes it a valuable model in practical applications where avoiding misclassification of non-conspiratorial content is crucial.

\begin{table}[H]
\centering
\caption{Model performance: NEC, NEC reducing FN, and NEC reducing FP. Lower is better.}
\begin{tabular}{l l c c c}
\toprule
Model & NEC & Reduce FN & Reduce FP \\
\midrule
Always Positive & 3.533 & 1.000 & 31.800 \\
Always Negative & 1.000 & 2.547 & 1.000 \\
Llama & 1.267 & \textbf{0.811} & 9.800 \\
Mistral & 3.267 & 2.736 & 23.000 \\
RoBERTa & \textbf{1.067} & 1.811 & \textbf{4.267} \\
\bottomrule
\end{tabular}
\end{table}

\begin{figure}[!t]
    \centering
    \subfloat{
        \includegraphics[width=0.3\textwidth]{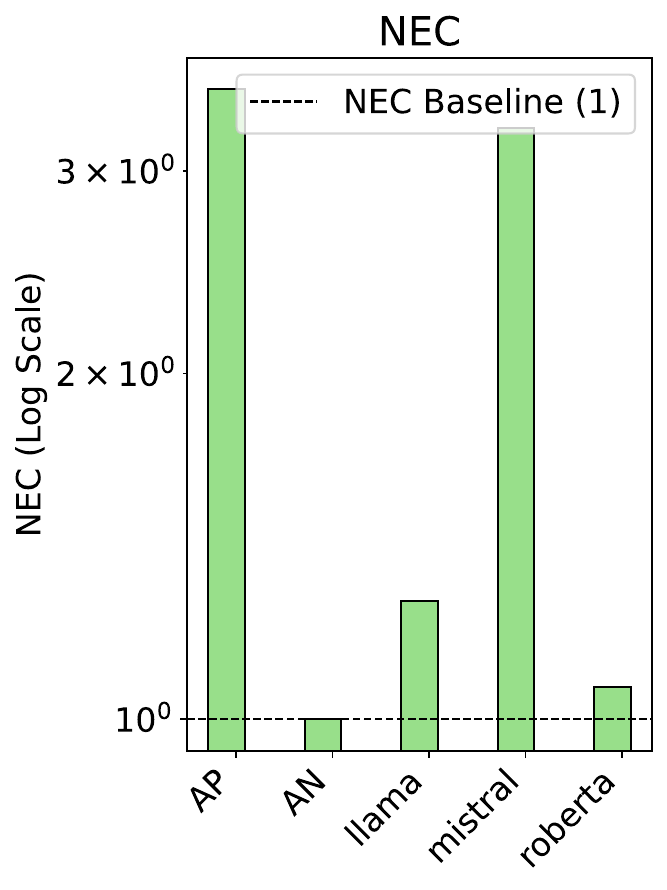}
        \label{fig:nec_itw}
    }
    \hfill
    \subfloat{
        \includegraphics[width=0.3\textwidth]{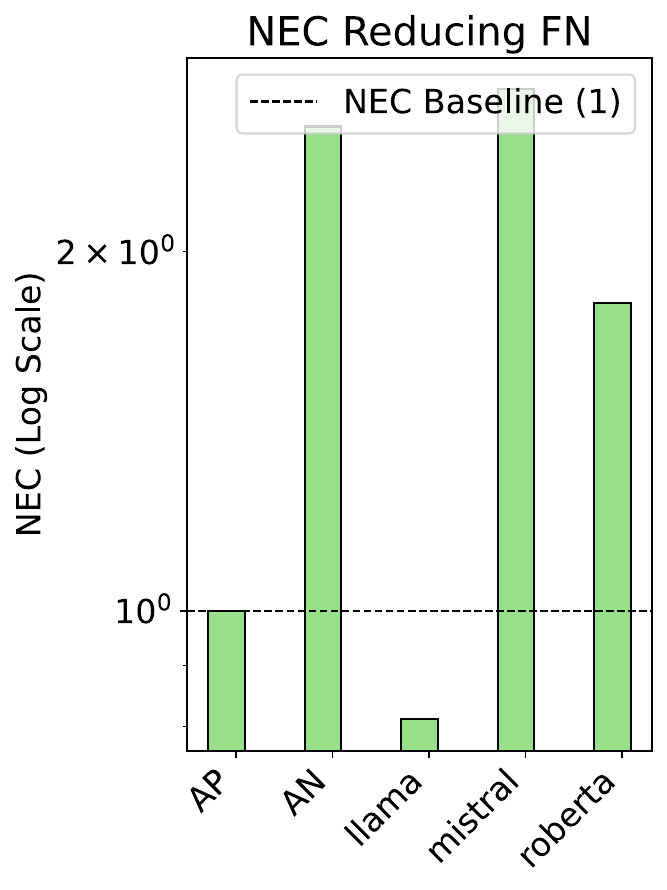}
        \label{fig:nec_reduce_fn_itw}
    }
    \hfill
    \subfloat{
        \includegraphics[width=0.3\textwidth]{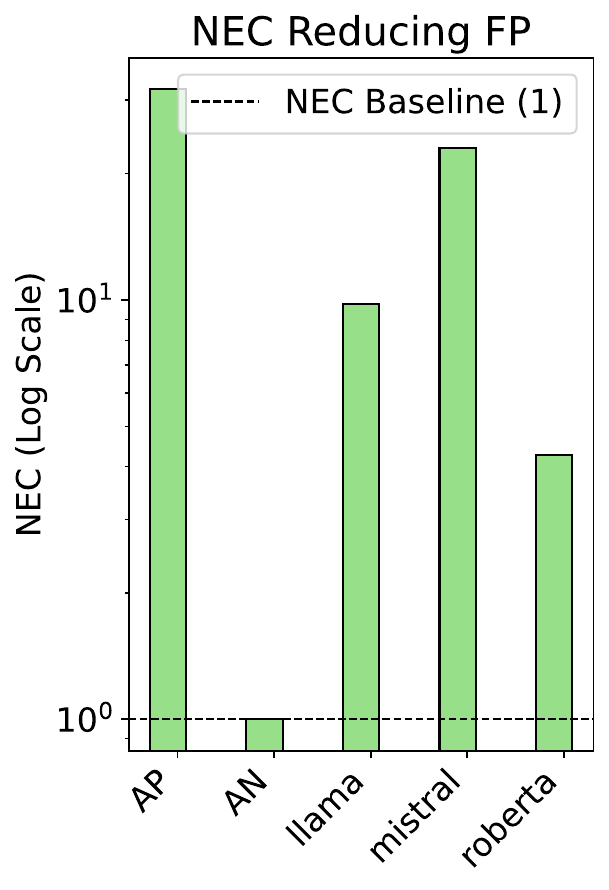}
        \label{fig:nec_reduce_fp_itw}
    }
    \caption{Comparison of NEC, NEC reducing FN, and NEC reducing FP across models.}
    \label{fig:NEC_itw}
\end{figure}

To further evaluate model performance, we analyzed the NEC scores across different classification settings.
The results indicate that RoBERTa performs best overall, reinforcing its effectiveness in reducing false positives while maintaining reasonable classification performance. LLaMA follows,  excelling in reducing false negatives due to its high recall, which allows it to detect more conspiracy content. These findings align with the precision-recall results, further supporting RoBERTa’s suitability for a conservative classification approach in real-world settings.

\section{Discussion}

\subsection{Contributions}


We constructed a multimodal dataset by enriching the YouNiCon dataset with additional metadata, video transcripts, and images, enabling the evaluation of both text-only and multimodal models.
The hydration pipeline used to enrich the datasets is available on \href{https://github.com/leoli51/youtube-conspiracy-detection}{GitHub}.
We conducted extensive zero-shot classification experiments with various open-weight LLMs—both text-only and multimodal—and found that while LLMs outperform their multimodal counterparts, fine-tuned models still achieve the best performance.
We benchmarked a range of LLMs from small (e.g., Mistral 7B) to large-scale (e.g., LLaMA 3.3 70B), providing empirical evidence that while larger models achieve the best performance, smaller models offer competitive performance with significantly lower computational costs.
We observed that LLMs tend to favor high recall, often at the cost of precision. This bias has practical implications for minimizing false positives.
We considered cost-sensitive evaluation metrics, particularly Normalized Expected Cost (NEC), to provide a nuanced analysis of model performance under varying classification priorities; this evaluation framework can be used to practical decision-making in real-world applications. 


\subsection{Implications}

The superior performance of fine-tuned models underscores the continued value of task-specific adaptation, especially when high precision is required for content moderation.
The limited effectiveness of multimodal models suggests that current visual inputs (e.g., thumbnails) offer minimal additional value for conspiracy detection, guiding future resource allocation.
The strong performance of smaller models like Mistral 7B demonstrates the feasibility of cost-efficient deployment, enabling scalable moderation in resource-constrained settings.
The observed trade-off between precision and recall in LLMs highlights the need for careful evaluation strategies in real-world deployment, especially where false positives carry social or platform risks.

\subsection{Limitations}

The reliance on zero-shot classification means that the LLM models were not provided with example cases, potentially limiting their ability to adapt to the specific characteristics of YouTube conspiracy narratives. 
The multimodal approach was restricted to static images (thumbnails), and did not incorporate full video information, limiting the assessment of visual misinformation strategies, furthermore the study assessed the performance of only one multimodal model.
The study primarily focused on English-language content, limiting generalizability to other languages where conspiracy discourse may differ.
We did not thoroughly explore the performance of the RoBERTa model, as it was primarily included as a fine-tuned baseline for comparison.
We found inconsistencies in the human-attributed labels, highlighting potential annotation biases.
Regarding the in-the-wild evaluation, our dataset was not a fully random sample of YouTube videos but was drawn from the YouTube-Commons dataset, which contains videos with Creative Commons licenses.
This led to a category distribution that differs from the natural composition of YouTube content, particularly with a higher representation of educational material.
While this dataset provided a valuable testing ground for model generalization, its category bias may impact conclusions about performance in truly unconstrained, real-world scenarios where misinformation may follow different distribution patterns.

\subsection{Future Work}

Future research can build upon this study by addressing the identified limitations and exploring new directions. Incorporating additional multimodal elements, such as optical character recognition (OCR) and deeper video frame analysis, to assess whether richer visual features improve classification. Investigating better fine-tune or few-shot learning approaches to refine model predictions without requiring extensive labeled training data. Expanding the study to non-English languages to evaluate cross-linguistic generalization and cultural variations in conspiracy narratives. Exploring different labeling strategies to reduce subjective biases in the human annotations. Exploring explainability techniques to provide interpretable classification rationales, improving transparency in automated misinformation detection. Furthermore, future work should consider alternative in-the-wild evaluation strategies that better reflect the organic distribution of conspiracy-related content on YouTube.

These future directions aim to improve the reliability and applicability of LLM-based misinformation detection while addressing gaps identified in this study.

\subsection{Ethical Aspects}

The detection of conspiracy content on social media platforms presents several ethical challenges that must be carefully managed. LLMs inherently carry biases from their training data, which can influence classification decisions, making fairness and the prevention of excessive flagging of specific viewpoints a critical concern. High false positive rates risk unjustified content removal, leading to potential censorship and suppression of legitimate discourse, necessitating a delicate balance between misinformation mitigation and freedom of expression. Transparency and accountability are essential in automated classification systems, requiring clear explanations of decisions to enable external audits and appeals. LLMs do not inherently provide this level of interpretability, making it challenging to fully understand their decision-making processes. Additionally, privacy concerns arise from the use of metadata and comment analysis, underscoring the need for strict ethical data practices. The deployment of misinformation detection models on platforms like YouTube carries broader societal implications, affecting public discourse and information accessibility. Therefore, continuous evaluation of these impacts is necessary to mitigate unintended harm. 

\subsection{Acknowledgements}
Francesco Pierri is partially supported by FAIR (Future Artificial Intelligence Research) project, funded by the NextGenerationEU program within the PNRR-PE-AI scheme (M4C2, Investment 1.3, Line on Artificial Intelligence).
The work in this paper was originally submitted as a Master Thesis titled: “Multimodal LLMs vs LLMs vs RoBERTa: Evaluating AI Performance in Detecting Conspiracies on YouTube” written by Leonardo La Rocca and supervised by Prof. Francesco Pierri and Francesco Corso.

\bibliographystyle{abbrvnat}
\bibliography{bibliography}

\appendix

\section{Qualitative Error Analysis}

In addition to computing classification metrics, we conducted a qualitative review of misclassified instances to better understand the nature of the errors made by each model. For each model-prompt combination, we selected three false positive and three false negative instances and manually reviewed their content. The goal of this analysis was to categorize the type of error and determine whether it was caused by:

\begin{itemize}
    \item A model mistake: the model incorrectly classified an instance due to limitations in its understanding or reasoning.
    \item A mislabeled instance: the ground truth label provided by the dataset was incorrect or ambiguous.
    \item A controversial case: the video contained elements that made classification inherently difficult, such as satire, ambiguous phrasing, or partial misinformation.
\end{itemize}

By systematically analyzing these cases, we aimed to identify patterns in classification errors, helping to assess whether certain types of content were more likely to be misclassified.




A significant number of mislabeled instances were identified, where the assigned ground-truth label did not align with the codebook definitions. This misalignment highlights inherent ambiguities in conspiracy content classification. Many videos contained elements attributable to both conspiracy and non-conspiracy narratives, making their categorization highly subjective. Annotators of the YouNICon dataset had to rely on limited metadata: title, description, tags, and the first 1,000 characters of the transcript, potentially omitting crucial contextual information presented deeper in the video.

A key difficulty in classification was the nuanced nature of conspiracy content. Some videos embedded conspiracy elements subtly, only introducing them after an extended runtime. Others presented content that was contextually dependent, making it difficult to determine their stance without full engagement. Furthermore, some videos lacked transcripts entirely and relied on on-screen text, which was unavailable to both our LLM-based classifiers and human annotators. These cases suggest that classification efforts could benefit from additional multimodal capabilities, such as optical character recognition (OCR) and deeper video analysis.

Among the most problematic cases were videos discussing religious beliefs intertwined with conspiracy narratives. Determining when religious claims cross into conspiracy territory is inherently subjective, as these perspectives are often framed as doctrinal rather than deceptive. This observation suggests that a rigid binary classification may not fully capture the nuances of conspiracy discourse. Instead, a multi-label approach could provide a more flexible framework, allowing for varying degrees of conspiracy-related content attribution.

Given the subjectivity and contextual dependence observed in our error analysis, alternative classification strategies should be explored. One possible approach is a multi-label classification scheme, where videos could be assigned multiple labels reflecting different aspects of their content. This would allow for a more granular understanding of borderline cases where conspiracy themes are present but not dominant.

Another promising direction is the integration of explainable AI (XAI) techniques. By leveraging model interpretability methods, we could provide justifications for classifications, increasing transparency in automated labeling decisions. Such an approach could mitigate concerns about subjectivity by exposing the decision-making process, enabling both human auditors and future annotation efforts to refine classification criteria.

In summary, our qualitative error analysis underscores the inherent difficulties in defining and detecting conspiracy content. Mislabeling often stemmed from contextual ambiguity, reliance on partial metadata, and the subtlety of conspiracy elements within broader narratives. These findings motivate the need for a more flexible and interpretable classification approach.

\section{In the Wild Dataset features}
\label{app:inw-dataset}
\begin{figure}[!t]
    \centering
    \includegraphics[width=\textwidth]{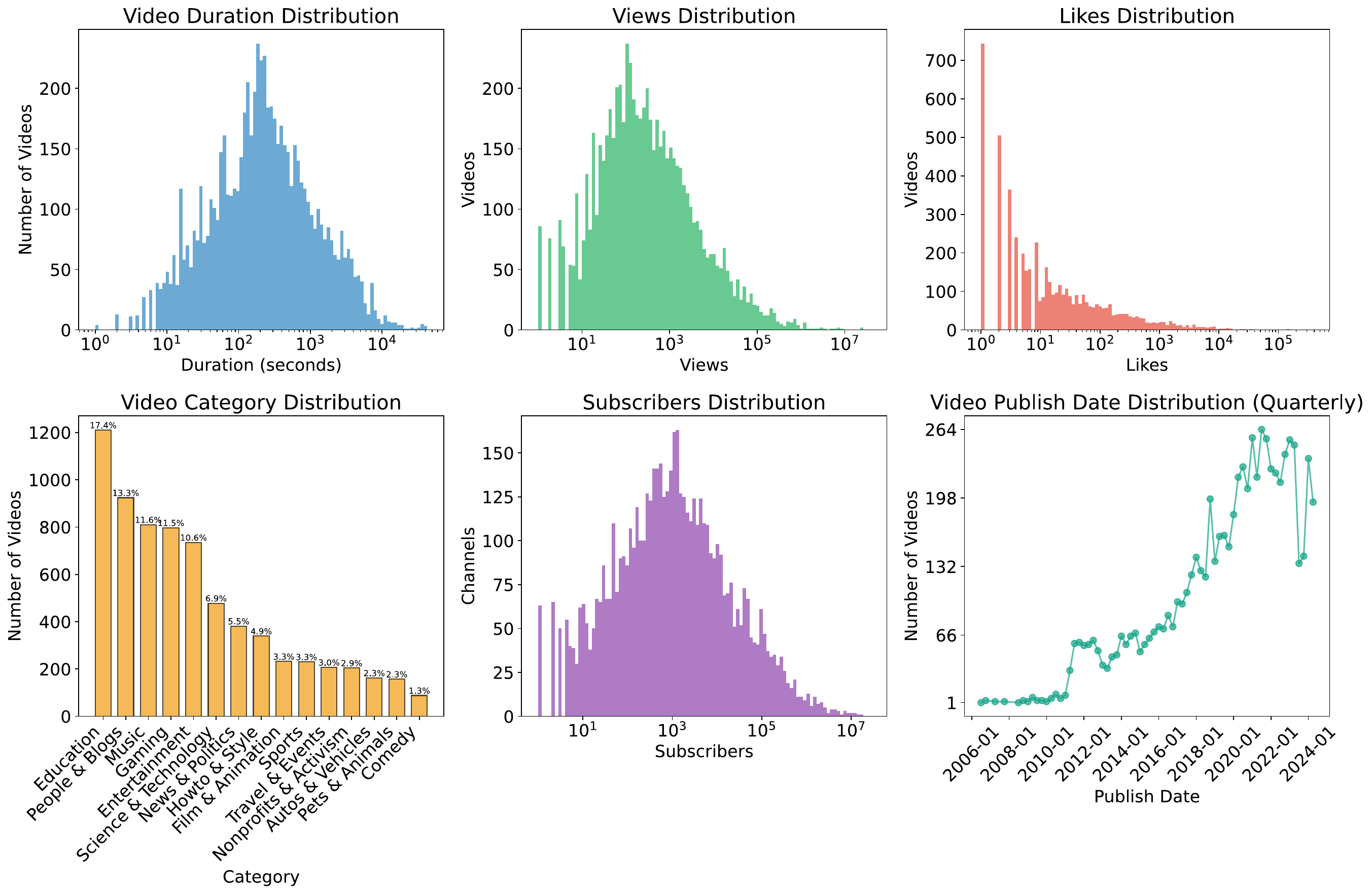}
    \caption{Distributions of various variables of the "in-the-wild" dataset.}
    \label{fig: Distributions of various variables of the "in-the-wild" dataset.}
\end{figure}

\begin{figure}[!t]
    \centering
     \includegraphics[width=\textwidth]{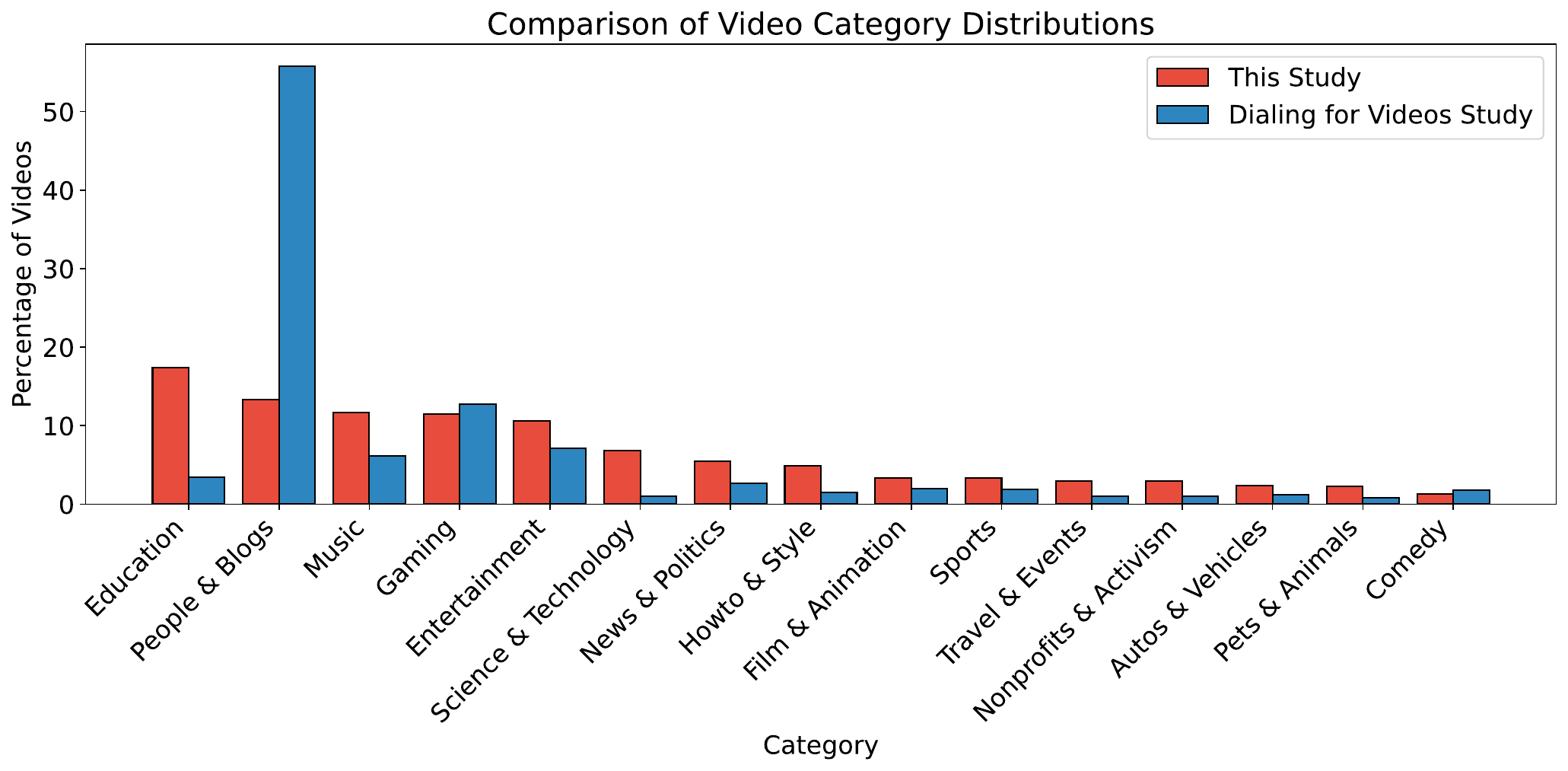}
    \caption{Comparison of video category distributions.}
    \label{fig: Comparison of video category distributions.}
\end{figure}

A key difference between Youtube Commons and the dataset collected by \citet{dialing_for_videos_2023} lies in the distribution of video categories.
As shown in Figure \ref{fig: Comparison of video category distributions.}, while the "dialing for videos" study reflects a stronger imbalance in the video categories, presenting a high number of "People and Blogs" videos, our dataset is notably more balanced across categories, with a particularly higher representation of educational videos. 
This can be attributed to the nature of YouTube-Commons, which consists of freely licensed Creative Commons content, often prevalent in educational resources. Although this dataset does not perfectly mirror the natural distribution of YouTube content, it still serves its purpose by enabling an assessment of model robustness across diverse content categories.

\section{Additional Tables}
\label{sec:additional tables}

\begin{table}[H]
\centering
\caption{Model performance: Precision and Recall in a balanced setting.}
\begin{tabular}{l l c c}
\toprule
Model & Prompt & Precision & Recall \\
\midrule
Roberta 3Epochs & none & - & - \\
ensemble-small-models & simple & 0.689 & 0.884 \\
ensemble-small-models & definition & 0.655 & 0.939 \\
google/gemma-1.1-7b-it & definition & 0.619 & 0.944 \\
google/gemma-1.1-7b-it & simple & 0.567 & 0.968 \\
meta-llama/Llama-3.1-8B-Instruct & definition & 0.660 & 0.914 \\
meta-llama/Llama-3.1-8B-Instruct & simple & 0.697 & 0.859 \\
meta-llama/Llama-3.3-70B-Instruct & simple & 0.703 & 0.920 \\
meta-llama/Llama-3.3-70B-Instruct & definition & 0.745 & 0.852 \\
mistralai/Mistral-7B-Instruct-v0.2 & definition & 0.662 & 0.887 \\
mistralai/Mistral-7B-Instruct-v0.2 & simple & \textbf{0.767} & 0.674 \\
meta-llama/Llama-3.2-11B-Vision-Instruct & definition & 0.541 & \textbf{0.949} \\
meta-llama/Llama-3.2-11B-Vision-Instruct & simple & 0.571 & 0.649 \\
\bottomrule
\end{tabular}
\end{table}

\begin{table}[H]
\centering
\caption{Model performance: NEC evaluated with a realistic class distribution.}
\begin{tabular}{l l c c c}
\toprule
Model & Prompt & NEC & Reduce FN & Reduce FP \\
\midrule
Always Positive & none                          & 19.000 & 2.111 & 171.000 \\
Always Negative & none                          & 1.000 & 1.000 & 1.000 \\
Roberta 3Epochs & none                          & \textbf{3.650} & \textbf{0.732} & \textbf{29.915} \\
ensemble-small-models & simple                  & 7.714 & 0.960 & 68.495 \\
ensemble-small-models & definition              & 9.467 & 1.106 & 84.716 \\
google/gemma-1.1-7b-it & definition             & 11.078 & 1.280 & 99.251 \\
google/gemma-1.1-7b-it & simple                 & 14.030 & 1.588 & 126.010 \\
meta-llama/Llama-3.1-8B-Instruct & definition   & 9.057 & 1.083 & 80.830 \\
meta-llama/Llama-3.1-8B-Instruct & simple       & 7.261 & 0.932 & 64.226 \\
meta-llama/Llama-3.3-70B-Instruct & simple      & 7.490 & 0.904 & 66.768 \\
meta-llama/Llama-3.3-70B-Instruct & definition  & 5.691 & 0.764 & 50.032 \\
mistralai/Mistral-7B-Instruct-v0.2 & definition & 8.732 & 1.070 & 77.686 \\
mistralai/Mistral-7B-Instruct-v0.2 & simple     & 4.236 & 0.760 & 35.519 \\
meta-llama/Llama-3.2-11B-Vision-Instruct & definition & 15.246 & 1.739 & 136.808 \\
meta-llama/Llama-3.2-11B-Vision-Instruct & simple & 9.568 & 1.375 & 83.308 \\
\bottomrule
\end{tabular}
\end{table}

\section{Additional Figures}
\label{sec:additional figures}

\begin{figure}[!t]
    \centering
     \includegraphics[width=\textwidth]{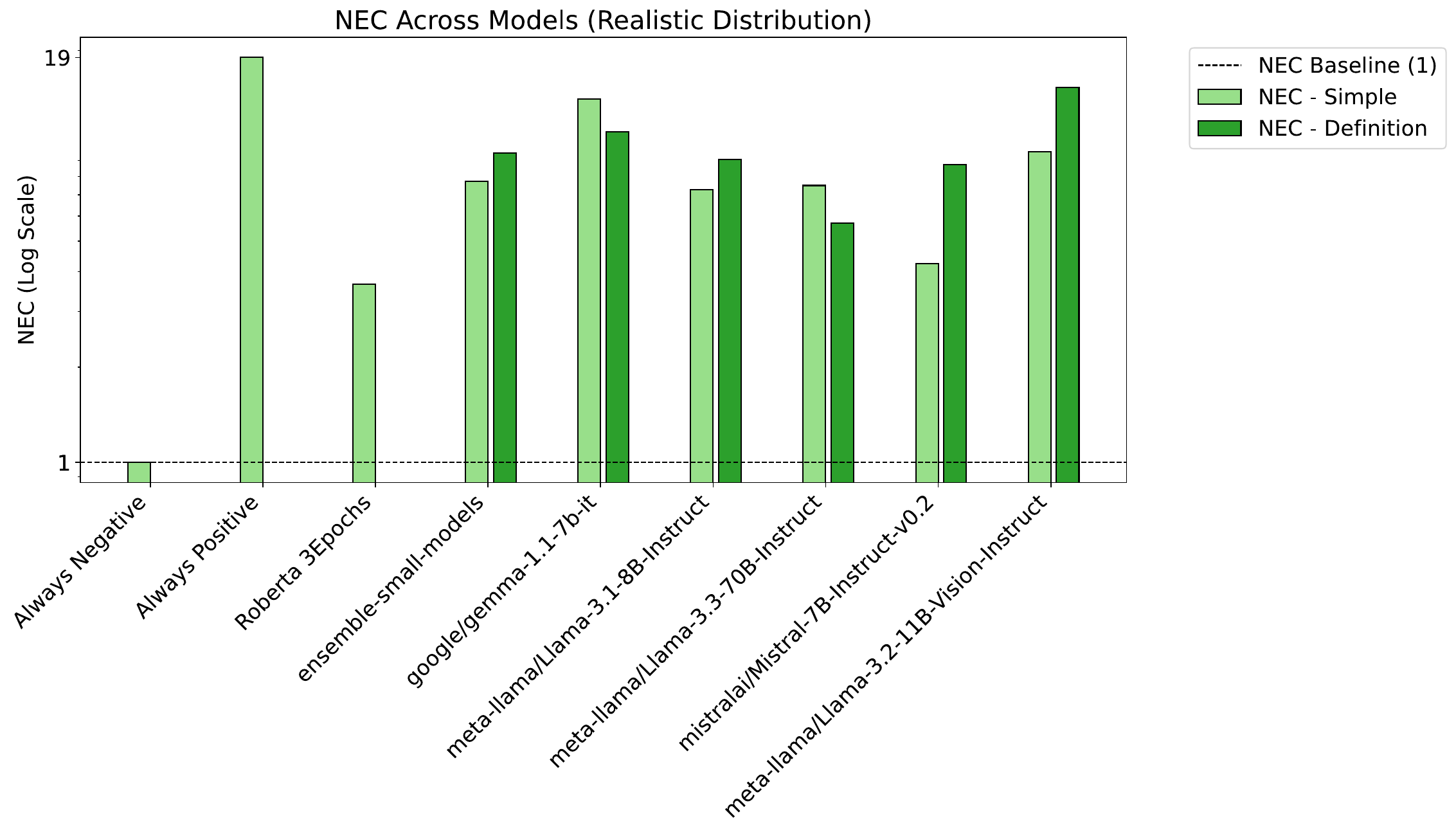}
    \caption{NEC across models and prompts (realistic distribution).}
    \label{fig:NEC across models (realistic distribution)}
\end{figure}

\begin{figure}[!t]
    \centering
     \includegraphics[width=\textwidth]{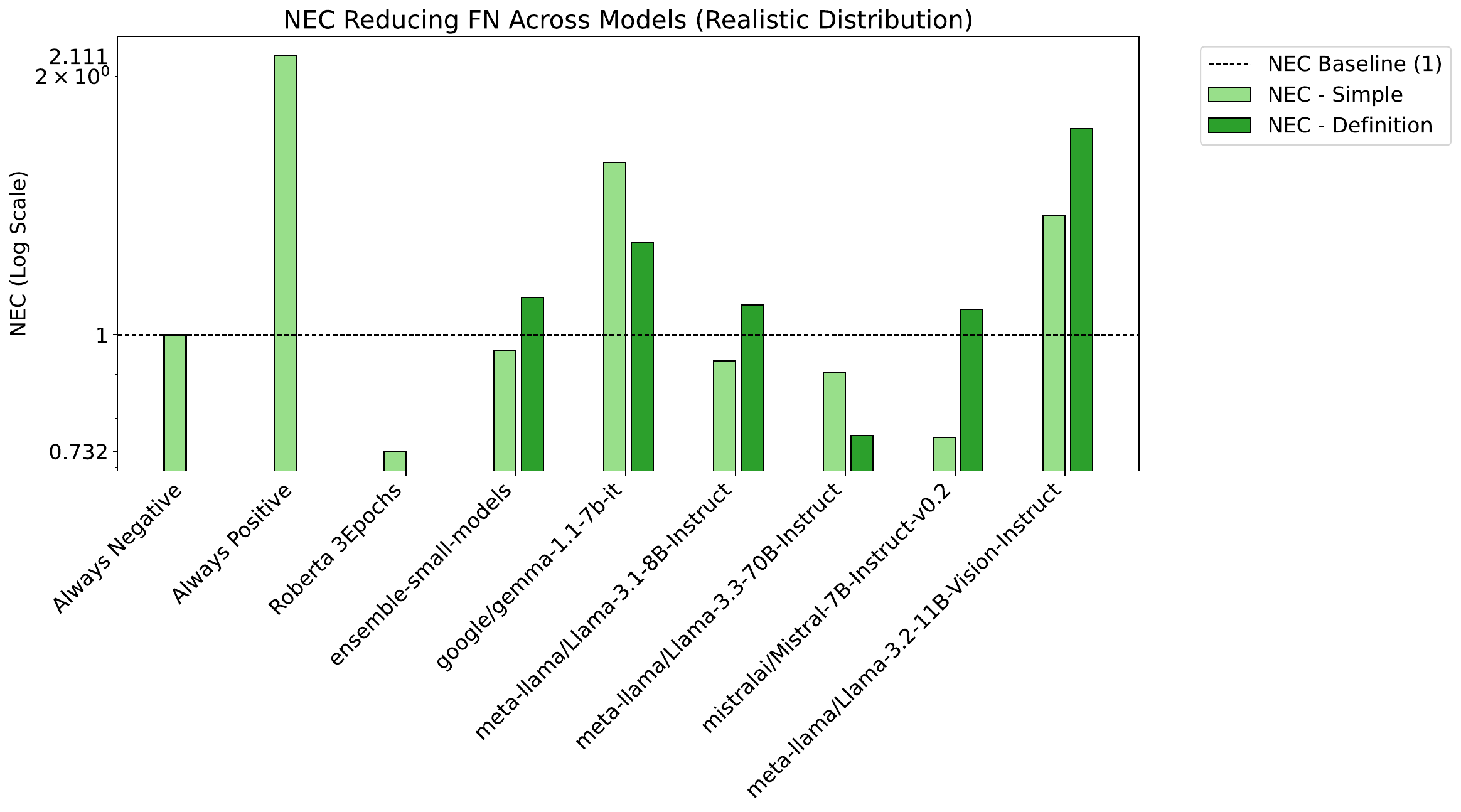}
    \caption{NEC reducing FN across models and prompts (realistic distribution).}
    \label{fig:NEC reducing FN across models (realistic distribution)}
\end{figure}

\begin{figure}[!t]
    \centering
     \includegraphics[width=\textwidth]{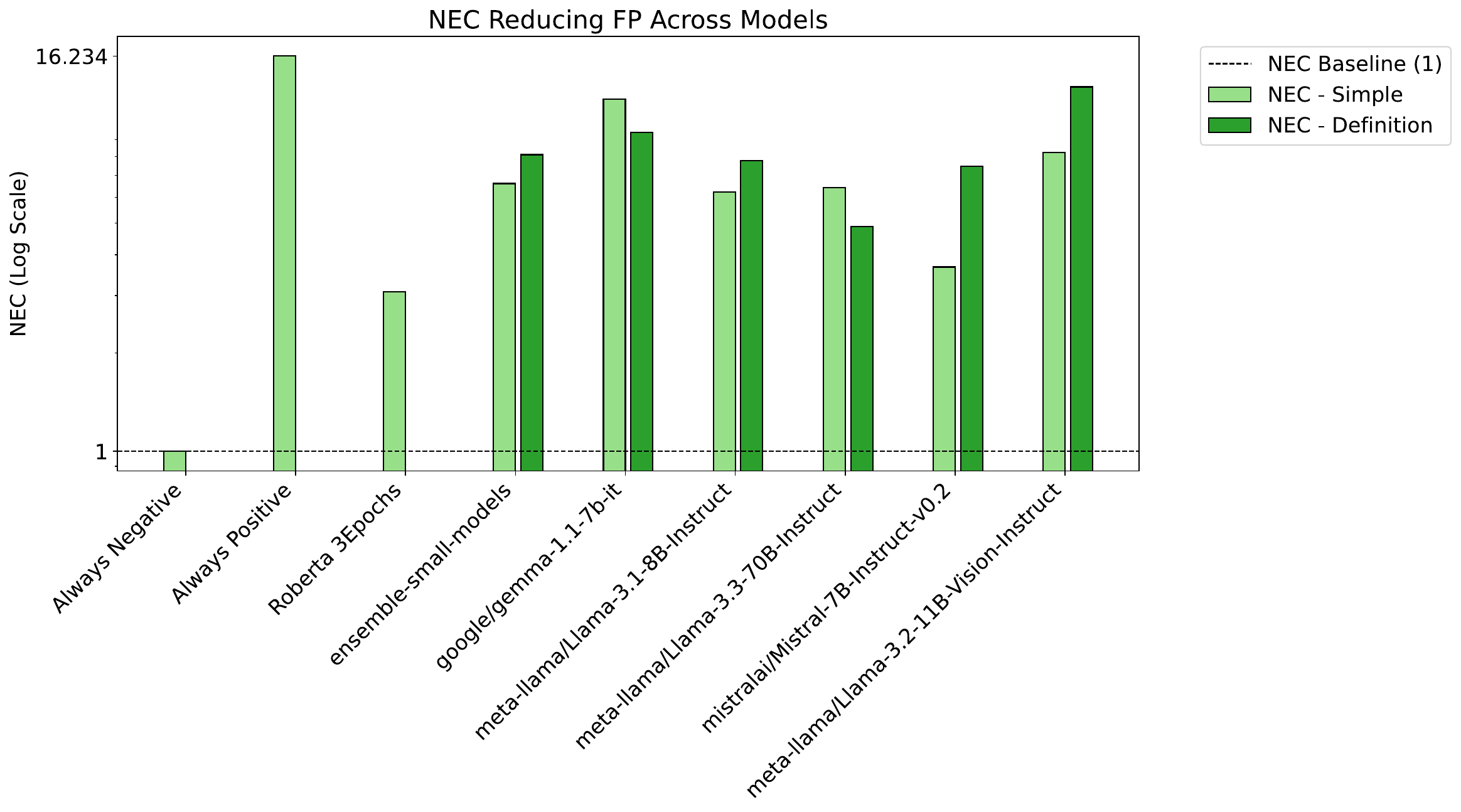}
    \caption{NEC reducing FP across models and prompts (realistic distribution).}
    \label{fig:NEC reducing FP across models (realistic distribution)}
\end{figure}

\end{document}